\documentclass[11pt]{article}

\usepackage[final]{acl}

\usepackage{times}
\usepackage{latexsym}

\usepackage[T1]{fontenc}

\usepackage[utf8]{inputenc}

\usepackage{microtype}

\usepackage{inconsolata}

\usepackage{graphicx}
\usepackage{amsmath}
\usepackage{amssymb}
\usepackage{multirow}
\usepackage{booktabs}
\usepackage{algorithm}
\usepackage{algpseudocode}
\usepackage{listings}
\usepackage{xcolor}
\usepackage[table]{xcolor}
\usepackage{tcolorbox}
\definecolor{MyDarkOrange}{RGB}{166, 82, 0}

\title{STAPO: Selective Trajectory-Aware Policy Optimization for \\ LLM Agent Training}

\author{
    Qiuyi Qi$^\spadesuit$$^\diamondsuit$\footnotemark[1],
    Tian Liang$^\spadesuit$$^\diamondsuit$\thanks{Q. Qi, T. Liang and M. Bao contributed equally to this work.},
    Mutian Bao$^\spadesuit$$^\diamondsuit$\footnotemark[1], 
    Jinjian Zhang$^\diamondsuit$,
    Dongnan Liu$^\diamondsuit$, \\
    \textbf{Wei Zhou}$^\diamondsuit$,
    \textbf{Linjian Mo}$^\diamondsuit$,
    \textbf{Ming Kong}$^\spadesuit$\footnotemark[2],
    \textbf{Jie Liu}$^\clubsuit$\footnotemark[2], \\
    \textbf{Feng Zhang}$^\spadesuit$,
    \textbf{Qiang Zhu}$^\spadesuit$\thanks{Q. Zhu, M. Kong and J. Liu are corresponding authors. Q. Zhu is with the College of Artificial Intelligence, Shanghai Institute for Advanced Study, Zhejiang University. M. Kong is with the School of Earth Sciences, Zhejiang University. J. Liu is with the Department of Computer Science, City University of Hong Kong.}\\
    $^\spadesuit$ Zhejiang University, 
    $^\diamondsuit$ Ant Group,
    $^\clubsuit$ City University of Hong Kong \\
    \texttt{
    \{qiqiuyi,zhuq\}@zju.edu.cn 
    }
}

\begin{document}
\maketitle
\begin{abstract}

Reinforcement Learning (RL) is the dominant paradigm for training Large Language Model (LLM) agents on long-horizon tasks. However, sparse and delayed rewards often lead to \textit{trajectory neglect}, in which agents lose focus on the task goal and interaction history at intermediate steps.
Prior work has explored step-level supervision using Shannon-entropy–based uncertainty signals, which conflate inherent state complexity with agent confidence and therefore provide unreliable estimates of decision reliability. To address this issue, we propose \textit{normalized entropy}, which measures confidence deviations relative to an agent’s average behavior under a given state, thereby strengthening the association between low-quality actions and trajectory neglect.
Building on this insight, we introduce \textbf{S}elective \textbf{T}rajectory-\textbf{A}ware \textbf{P}olicy \textbf{O}ptimization (STAPO), a hierarchical group-based RL framework. STAPO leverages normalized entropy to locate outlier steps associated with trajectory neglect and optimizes them via a joint mechanism of \textit{trajectory-aware reward} and \textit{trajectory-independent penalty}, enhancing trajectory awareness while preserving training stability.
Extensive experiments on ALFWorld, WebShop, and Search-Augmented QA demonstrate that STAPO achieves state-of-the-art performance while substantially alleviating trajectory neglect, validating its effectiveness and robustness for agentic tasks.
\end{abstract}

\begin{figure}[t]
\centering
  \includegraphics[width=\columnwidth]{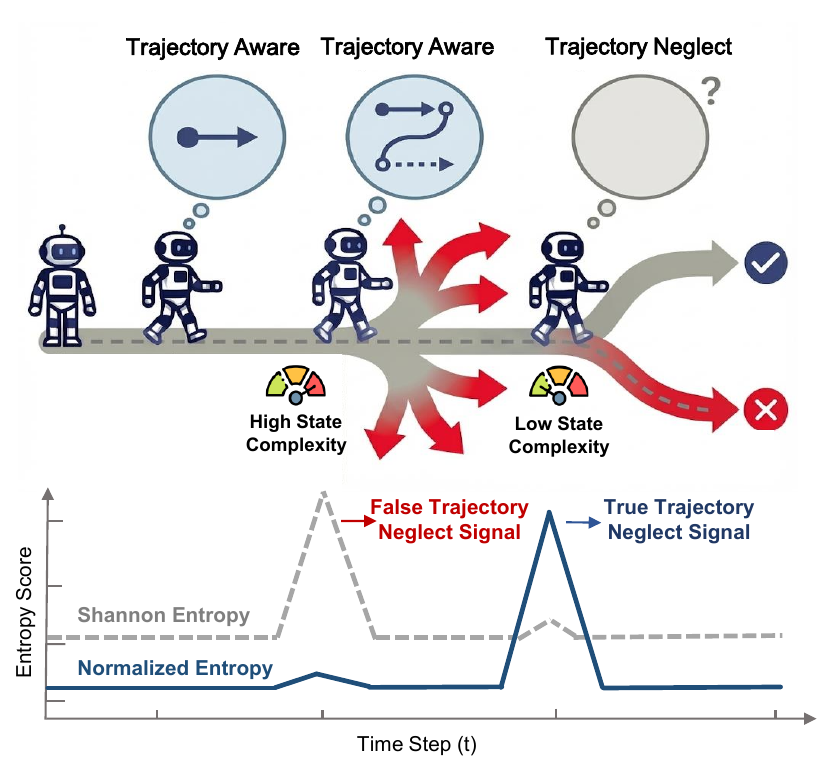}
  \caption{Illustration of the differences between \textit{Shannon entropy} and \textit{normalized entropy} in locating trajectory neglect.}
  \label{fig:fig1}
\end{figure}

\section{Introduction}
Large Language Models (LLMs) have evolved from static text generators to autonomous agents capable of executing complex, long-horizon tasks, such as embodied control~\citep{shridhar2020alfworld,li2024embodied} and web navigation~\citep{furuta2024multimodal,zheng2024gpt,gou2025navigating}.

To enable long-horizon reasoning in LLM agents, recent work has increasingly adopted {Reinforcement Learning (RL)} for post-training. In particular, group-based RL algorithms such as RLOO~\citep{kool2019buy,ahmadian2024back} and GRPO~\citep{shao2024deepseekmath} estimate advantages via group sampling, avoiding explicit critic training while improving scalability and training stability.
However, training agents for long-horizon tasks remains challenging due to the \textit{temporal credit assignment} problem induced by sparse and delayed rewards. As illustrated in Figure~\ref{fig:fig1}, agents often exhibit \textit{trajectory neglect}, where the model loses focus on the task goal and interaction history during extended interaction sequences, leading to low-quality actions at certain intermediate steps (see Appendix~\ref{sec:CSTN} for detailed case studies).

While recent adaptations like GiGPO~\citep{feng2025group} and RLVMR~\citep{zhang2025rlvmr} attempt to mitigate the credit assignment issue by achieving fine-grained step-level advantage estimation, they typically optimize all steps indiscriminately, lacking a selective mechanism to precisely locate and address these outlier steps caused by trajectory neglect. Furthermore, several works have explored leveraging internal uncertainty signals to assist training, including entropy-based regularization and supervision strategies~\citep{xu2025epo,cui2025entropy}. However, these approaches predominantly rely on \textit{Shannon entropy} computed from step-level action distributions, which conflates inherent state complexity with agent confidence. As shown in Figure~\ref{fig:fig1}, a high-entropy step may simply reflect a genuinely complex state with numerous plausible actions, rather than a lack of confidence. Consequently, raw Shannon entropy provides an unreliable signal for locating outlier steps associated with trajectory neglect (as empirically illustrated in Appendix~\ref{sec:appendix_CSNAE}).

To address these limitations, we first propose a novel metric: \textit{normalized entropy}. By aggregating actions from the same state across sampled trajectories, normalized entropy measures uncertainty relative to the state average, effectively decoupling inherent state complexity from model confidence. Our empirical analysis shows that outliers located by normalized entropy consistently correlate with trajectory neglect, which we empirically find to be strongly associated with low-quality actions.

Building on this insight, we introduce \textbf{S}elective \textbf{T}rajectory-\textbf{A}ware \textbf{P}olicy \textbf{O}ptimization (STAPO), a hierarchical group-based RL framework. STAPO operates in two phases: (1) \textbf{Outlier Localization}, which dynamically locates outlier steps using normalized entropy statistics; and (2) \textbf{Selective Optimization}, which targets these steps associated with trajectory neglect. Specifically, STAPO incorporates a \textit{trajectory-aware reward} to encourage stronger focus on trajectories, balanced by a \textit{trajectory-independent penalty} to mitigate the risk of reward hacking.

Extensive experiments on representative agentic benchmarks, including ALFWorld~\citep{shridhar2020alfworld}, WebShop~\citep{yao2022webshop}, and Search-Augmented QA, demonstrate that STAPO consistently outperforms strong baselines such as GiGPO~\citep{feng2025group} and EMPG~\citep{wang2025harnessing}, while attribution analysis further confirms its effectiveness in alleviating trajectory neglect.

Our contributions are summarized as follows:
\begin{itemize}
    \item We propose \textit{normalized entropy} to decouple state complexity from agent confidence and effectively locate low-quality actions and trajectory neglect.
    \item We introduce \textit{STAPO}, a hierarchical group-based RL framework that integrates outlier localization with a joint optimization mechanism of trajectory-aware reward and trajectory-independent penalty.
    \item Extensive experiments demonstrate that STAPO achieves superior performance and robustness across diverse agentic tasks, while substantially mitigating trajectory neglect.
\end{itemize}

\section{Related Work}

\subsection{LLM Agents}
Large Language Models (LLMs) have increasingly been developed into autonomous agents capable of perceiving, reasoning, and acting in dynamic environments~\citep{zhang2023you,wang2023voyager,brohan2023rt}. Early approaches primarily relied on prompt-based methods, such as ReAct~\citep{yao2023react} and Reflexion~\citep{shinn2024reflexion}, as well as enhanced memory, retrieval mechanisms~\citep{wang2024mobile,tan2024cradle}, and integration with external tools~\citep{schick2023toolformer,xie2024osworld,zhang2024ufo}. More recent work has shifted toward adapting model parameters through supervised fine-tuning (SFT)~\citep{zhang2023you} or reinforcement learning (RL)~\citep{sutton2018reinforcement}, enabling agents to learn directly from environment interactions.

\subsection{Group-based Policy Optimization}
To avoid the high overhead and instability associated with training independent critic models, group-based RL methods have emerged as promising alternatives to traditional actor-critic methods (like PPO~\citep{schulman2017proximal}). These algorithms, such as GRPO~\citep{shao2024deepseekmath}, RLOO~\citep{kool2019buy,ahmadian2024back}, and DAPO~\citep{yu2025dapo} leverage a group of samples from the same query and estimate the advantages accordingly. Recently, GiGPO~\citep{feng2025group}, RLVMR~\citep{zhang2025rlvmr}, and Tree-GRPO~\citep{ji2025tree} further advance this paradigm by introducing hierarchical grouping structures to achieve fine-grained step-level advantage estimation without relying on auxiliary models.
However, they optimize all steps indiscriminately, lacking a selective mechanism to precisely locate and address outlier steps caused by trajectory neglect.

\subsection{Entropy-based Reinforcement Learning} 
To overcome the challenges of sparse external rewards, recent works have explored using internal feedback (like model uncertainty, frequently quantified by entropy) to assist training~\citep{agarwal2025unreasonable}. However, most of them rely on raw Shannon entropy, which conflates inherent state complexity with agent confidence, and primarily utilize it as a proxy for regularization or re-weighting. For instance, EPO~\citep{xu2025epo} stabilizes exploration via entropy smoothing regularization; EMPG~\citep{wang2025harnessing} modulates the advantage term using entropy weights; and \citet{cui2025entropy} leverages entropy to truncate gradients for specific tokens. In contrast, we propose normalized entropy to decouple inherent state complexity from agent confidence, employing it to precisely locate outlier steps for selective trajectory-aware optimization.

\section{Empirical Analysis}
\label{sec:ps}
In this section, we formalize the task definitions, detail our statistical strategy for outlier localization based on a novel normalized entropy metric, and empirically validate its effectiveness in locating low-quality actions and trajectory neglect.

\subsection{Task Definitions}
\label{sec:td}
Formally, we consider a general setting where an LLM agent interacts with an environment in discrete time steps $t=1, 2, \dots, T$. The initial input instruction $x$ consists of a \textbf{system prompt} $p$ (defining general rules and roles) and a specific \textbf{task goal} $g$. At each step $t$, the agent receives a \textbf{current observation} $s_t$ and maintains an \textbf{interaction history} $h_t = (s_1, a_1, \dots, s_{t-1}, a_{t-1})$. The complete prompt at step $t$ is denoted as $c_t = (p, g, h_t, s_t)$, conditioned on which the model policy $\pi_\theta$ generates an action $a_t$.

To analyze agent behavior, we sample a group of $N$ full trajectories $\{\tau_1, \tau_2, \dots, \tau_N\}$ for each instruction $x$. We define an \textit{anchor state} $\tilde{s}$ as the canonical state identity used for grouping semantically equivalent step contexts within the current batch (exact-match keys for structured observations and similarity-based keys for free-text observations). Within a single training epoch, each sampled step is assigned to \emph{exactly one} anchor state via hard clustering (exact match or greedy maximum-similarity assignment), so no step simultaneously belongs to multiple groups. We then utilize the \textit{anchor state grouping} mechanism~\citep{feng2025group} to aggregate actions from the same state across sampled trajectories. Let $\mathcal{G}^S(\tilde{s}) = \{(a_t^{(i)}, R_t^{(i)}) \mid s_t^{(i)} = \tilde{s}, 1 \le i \le N, 1 \le t \le T\}$ denote the step-level group for an anchor state $\tilde{s}$, where $a_t^{(i)}$ is the action taken at step $t$ in trajectory $\tau_i$, and $R_t^{(i)}$ is the cumulative reward starting from step $t$ (back-calculated from the outcome reward $R(\tau_i)$).

\begin{figure}[t]
\centering
  \includegraphics[width=\columnwidth]{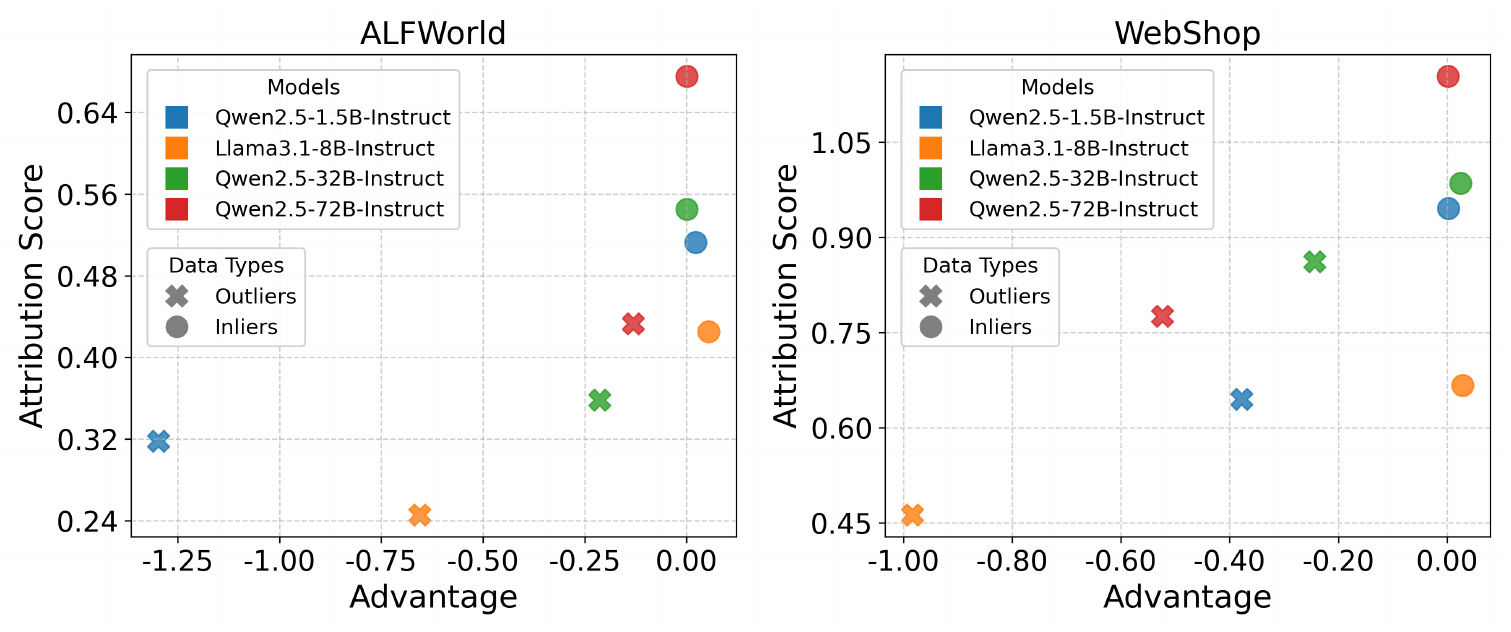}
  \caption{Empirical analysis on ALFWorld and WebShop. Markers denote the average step-level advantage and attribution score for outliers and inliers across various models.}
  \label{fig:fig_ea}
\end{figure}

\subsection{Outlier Localization}
\label{sec:ol}
To precisely locate outliers in model confidence, we implement a statistical strategy as follows:

\paragraph{Normalized Entropy Calculation.}
We first formalize the step-level Shannon entropy $H$. For an action $a_t^{(i)}$ consisting of a sequence of $M$ tokens $(w_1, \dots, w_M)$, $H(a_t^{(i)})$ is calculated as the average entropy over the generated tokens:
\begin{equation}
\small
H(a_t^{(i)}) = - \frac{1}{M} \sum_{k=1}^{M} \sum_{v \in \mathcal{V}} p(v \mid c_t^{(i)}, w_{<k}) \log p(v \mid c_t^{(i)}, w_{<k})
\label{eq:AE}
\end{equation}

where $p(v | c_t^{(i)}, w_{<k})$ is the probability of token $v$ from the vocabulary $\mathcal{V}$.

To decouple inherent state complexity from agent confidence, we propose normalized entropy $H_{n}$, which measures the deviation of $H$ from the state-specific baseline. This metric facilitates unbiased comparisons of confidence across diverse states in multi-turn interactions. We perform intra-group normalization on $H$ based on the statistics of the anchor state group $\mathcal{G}^S(\tilde{s})$:
\begin{equation}
\small
    H_{n}(a_t^{(i)}) = \frac{H(a_t^{(i)}) - \mu(\{H(a_t^{(j)}) \mid (a_t^{(j)}, \cdot) \in \mathcal{G}^S(\tilde{s})\})}{\sigma(\{H(a_t^{(j)}) \mid (a_t^{(j)}, \cdot) \in \mathcal{G}^S(\tilde{s})\}) + \epsilon}
    \label{eq:RE}
\end{equation}
where $\mu$ and $\sigma$ denote the mean and standard deviation respectively, and $\epsilon$ is a small constant added for numerical stability.

\begin{figure*}[t]
\centering
    \includegraphics[width=\linewidth]{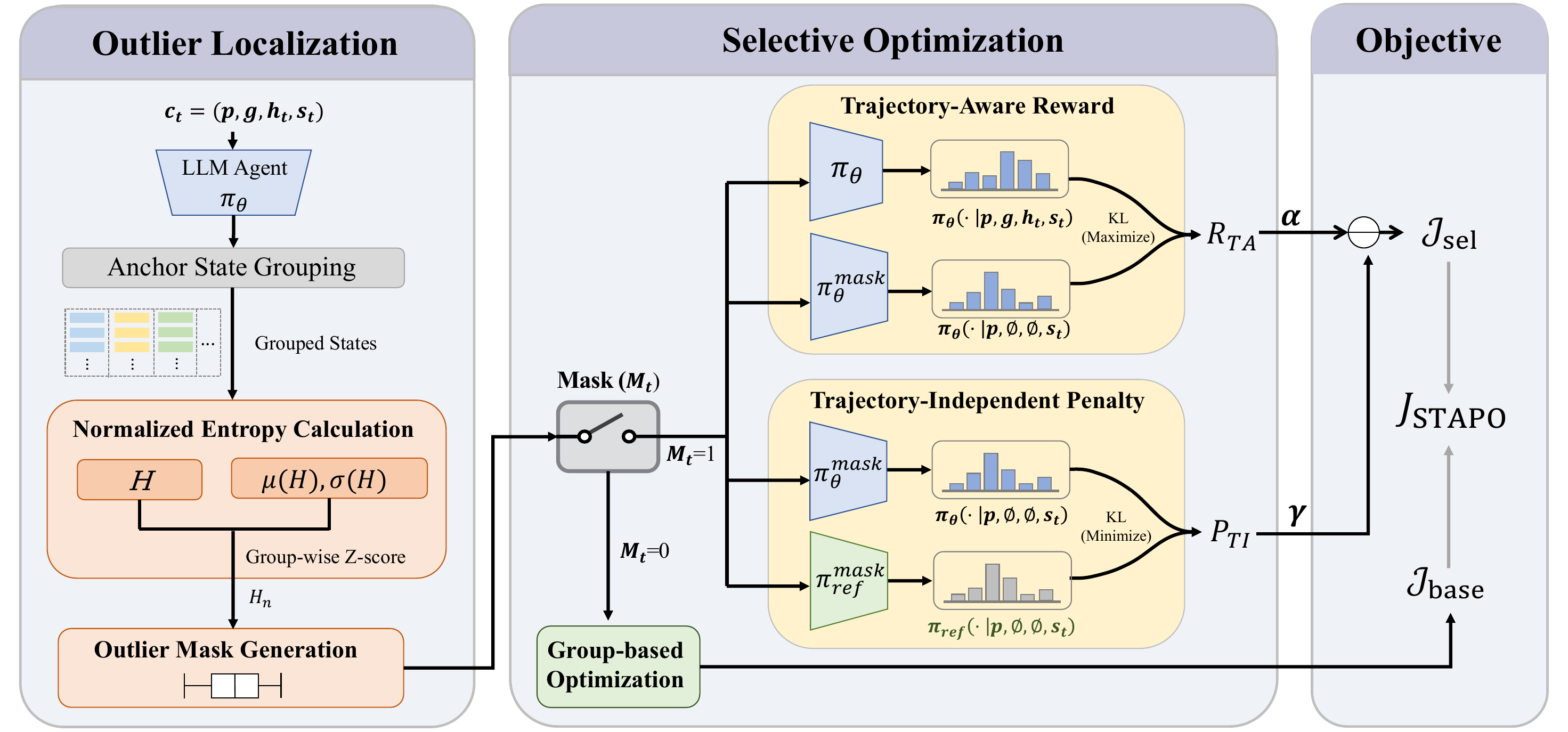}
    \caption{The framework of STAPO. It dynamically locates outlier steps utilizing normalized entropy statistics and optimizes them via a joint mechanism of trajectory-aware reward and trajectory-independent penalty.}
    
    \label{fig:fig2}
\end{figure*}

\paragraph{Outlier Mask Generation.}
Using $H_{n}$, we employ the Interquartile Range (IQR) method to dynamically locate outliers without manual thresholding. Let $\mathcal{R} = [Q_1 - \lambda \cdot \text{IQR},\, Q_3 + \lambda \cdot \text{IQR}]$ denote the inlier interval, where $Q_1$ and $Q_3$ are the first and third quartiles of $H_{n}$ in the current training batch, $\text{IQR} = Q_3 - Q_1$, and $\lambda$ is the IQR coefficient, typically set to $1.5$. We then define an indicator mask $M_t \in \{0, 1\}$ for each step:
\begin{equation}
\small
    M_t(a_t^{(i)}) = \mathbb{I}\!\left(H_{n}(a_t^{(i)}) \notin \mathcal{R}\right) \label{eq:IM}
\end{equation}
Steps with $M_t=1$ are flagged as outliers, while others are treated as inliers. Appendix~\ref{sec:appendix_pseudocode} lists pseudocode for outlier localization.

\subsection{Characterization of Outliers}
To validate the effectiveness of $H_{n}$ and investigate the nature of outliers, we conduct an empirical analysis on the test sets of ALFWorld~\citep{shridhar2020alfworld} and WebShop~\citep{yao2022webshop}. We visualize the relationship between action quality (measured by step-level advantage, calculated via the same intra-group normalization on $R_t^{(i)}$) and trajectory focus (measured by attribution score, detailed in Appendix~\ref{sec:appendix_DEA}) for both outliers and inliers. As shown in Figure~\ref{fig:fig_ea}:
\begin{itemize}
    \item \textbf{Outliers indicate low-quality actions.} Outliers (marked with `$\times$') are consistently clustered in the negative region of the Advantage axis. In contrast, inliers (marked with `$\bullet$') are positioned significantly to the right, exhibiting higher advantages. These results indicate that outliers tend to correspond to low-quality actions.
    \item \textbf{Outliers indicate trajectory neglect.} On the Attribution Score axis, outliers consistently reside in the lower region compared to their corresponding inliers. This indicates that at outlier steps, the model's focus on the trajectory is significantly weaker.
\end{itemize}
In summary, empirical results reveal that outliers located by normalized entropy serve as a reliable indicator of trajectory neglect, which is a primary cause of low-quality actions in long-horizon tasks.

\section{Methodology}
To address the challenge of trajectory neglect identified in Section~\ref{sec:ps}, we propose \textbf{S}elective \textbf{T}rajectory-\textbf{A}ware \textbf{P}olicy \textbf{O}ptimization (STAPO), a hierarchical group-based RL framework.

\subsection{Outlier Localization Integration}

As illustrated in the {left panel of Figure~\ref{fig:fig2}}, STAPO integrates the localization mechanism validated in Section~\ref{sec:ol} directly into the training loop. Specifically, for each step, we calculate the normalized entropy $H_n$ (Equation~\ref{eq:RE}) and derive the indicator mask $M_t$ (Equation~\ref{eq:IM}). This mask functions as a {dynamic gating mechanism}: it filters the training stream so that only steps flagged as outliers ($M_t=1$) trigger the {selective optimization} module, while inliers ($M_t=0$) are optimized solely using the base objective.

\subsection{Selective Optimization}

For the located outlier steps, we introduce a joint optimization mechanism, depicted in the {center panel of Figure~\ref{fig:fig2}}, composed of a trajectory-aware reward and a trajectory-independent penalty.

\paragraph{Trajectory-Aware Reward.} 
To encourage the model to focus on the trajectory when responding, we propose a trajectory-aware reward. We define a masking operation $(\cdot)_{\setminus \mathcal{T}}$ that strips the goal and history from the complete prompt $c_t^{(i)}$, yielding a trajectory-blind prompt $(c_t^{(i)})_{\setminus \mathcal{T}} = (p, \emptyset, \emptyset, s_t^{(i)})$. We then calculate the KL divergence between the model's output log-probabilities under these two conditions (denoted as $\pi_\theta$ and $\pi_\theta^{mask}$ in Figure~\ref{fig:fig2}):

\begin{equation}
\small
    R_{TA}(c_t^{(i)}, a_t^{(i)}) = \mathbb{D}_{KL}[ \pi_\theta(a_t^{(i)} | c_t^{(i)}) \parallel \pi_\theta(a_t^{(i)} | (c_t^{(i)})_{\setminus \mathcal{T}})]
    \label{eq:TAR}
\end{equation}

Conceptually, $R_{TA}$ measures how much the agent’s action distribution depends on the full trajectory context. A higher $R_{TA}$ implies that the agent is actively leveraging the trajectory information to shape its output distribution.

\paragraph{Trajectory-Independent Penalty.} 
To mitigate the risk of reward hacking on $R_{TA}$ at outlier steps, we introduce a trajectory-independent penalty. Unlike the standard global KL penalty which operates on the complete prompt $c_t^{(i)}$, this penalty specifically constrains the model's behavior on the masked prompt $(c_t^{(i)})_{\setminus \mathcal{T}}$ to remain close to the reference policy $\pi_{ref}$ (denoted as $\pi_\theta^{mask}$ and $\pi_{ref}^{mask}$ in Figure~\ref{fig:fig2}):

\begin{equation}
\scriptsize
    P_{TI}(c_t^{(i)}, a_t^{(i)}) = \mathbb{D}_{KL}[ \pi_\theta(a_t^{(i)} | (c_t^{(i)})_{\setminus \mathcal{T}}) \parallel \pi_{ref}(a_t^{(i)} | (c_t^{(i)})_{\setminus \mathcal{T}})]
    \label{eq:TIR}
\end{equation}

This constraint ensures that the increase in $R_{TA}$ stems from the model's enhanced focus on the trajectory, rather than from mode collapse.

\subsection{STAPO Objective}

As illustrated in the {right panel of Figure~\ref{fig:fig2}}, the STAPO framework integrates these selective terms into a group-based RL objective. Inspired by GiGPO~\citep{feng2025group}, we adopt a two-level advantage formulation:

\paragraph{Advantage Estimation.} 
First, the episode-level advantage $A^E$ evaluates the overall trajectory quality. For trajectory $\tau_i$, it is calculated via intra-group normalization of outcome rewards:

\begin{equation}
\small
    A^E(\tau_i) = \frac{R(\tau_i) - \mu(\{R(\tau_j)\}_{j=1}^N)}{\sigma(\{R(\tau_j)\}_{j=1}^N) + \epsilon}
    \label{eq:ELA}
\end{equation}

Second, the step-level advantage $A^S$ provides fine-grained credit assignment. Utilizing the anchor state group $\mathcal{G}^S(\tilde{s})$ defined in Section~\ref{sec:td}, we calculate the advantage of action $a_t^{(i)}$ relative to the average performance in that state:

\begin{equation}
\small
    A^S(a_t^{(i)}) = \frac{R_t^{(i)} - \mu(\{R_t^{(j)} \mid (a_t^{(j)}, R_t^{(j)}) \in \mathcal{G}^S(\tilde{s})\})}{\sigma(\{R_t^{(j)} \mid (a_t^{(j)}, R_t^{(j)}) \in \mathcal{G}^S(\tilde{s})\}) + \epsilon}
    \label{eq:SLA}
\end{equation}

The overall advantage $A$ combines these two terms: $A(a_t^{(i)}) = A^E(\tau_i) + \omega A^S(a_t^{(i)})$, where $\omega$ is a weighting coefficient.

\paragraph{Optimization Objective.}
Incorporating the estimated advantage and the selective optimization terms, 
the final clipped policy optimization objective for STAPO is:
\begin{equation}
\scriptsize
    {J}_{\mathrm{STAPO}}(\theta) = \mathbb{E}_{\substack{x \sim p(X) \\ \{\tau_i\}_{i=1}^N \sim \pi_{\theta_{\text{old}}}}} 
    \left[ \frac{1}{NT} \sum_{i=1}^{N} \sum_{t=1}^{T} 
    \bigl( \mathcal{J}_{\mathrm{base}}^{(i,t)} + \mathcal{J}_{\mathrm{sel}}^{(i,t)} \bigr) \right]
    \label{eq:objective}
\end{equation}
where $p(X)$ is the task distribution and $\rho_\theta(a_t^{(i)})$ is the importance 
sampling ratio. The two components are defined as:
\begin{equation}
\scriptsize
    \begin{split}
        \mathcal{J}_{\mathrm{base}}^{(i,t)} &= \min\bigl[ \rho_\theta(a_t^{(i)}) A(a_t^{(i)}),\, 
            \text{clip}\bigl(\rho_\theta(a_t^{(i)}), 1 \pm \epsilon\bigr) A(a_t^{(i)})\bigr] \\
        &\quad - \beta \mathbb{D}_{KL}[\pi_{\theta} \parallel \pi_{ref}]
    \end{split}
    \label{eq:jbase}
\end{equation}
\begin{equation}
\scriptsize
    \mathcal{J}_{\mathrm{sel}}^{(i,t)} = M_t \cdot \bigl( \alpha R_{TA}(c_t^{(i)}, a_t^{(i)}) 
        - \gamma P_{TI}(c_t^{(i)}, a_t^{(i)}) \bigr)
    \label{eq:jsel}
\end{equation}
where $\mathcal{J}_{\mathrm{base}}^{(i,t)}$ represents the base group-based RL objective, 
$\mathcal{J}_{\mathrm{sel}}^{(i,t)}$ denotes our proposed selective component, 
and $\alpha$, $\beta$, $\gamma$ are weighting coefficients.

\section{Experiments}

\begin{table*}[t]
\centering
\small 
\resizebox{\textwidth}{!}{
\begin{tabular}{llccccccc|cc}
\toprule

\multirow{2}{*}{Type} & \multirow{2}{*}{Method} & \multicolumn{7}{c|}{\textbf{ALFWorld}} & \multicolumn{2}{c}{\textbf{WebShop}} \\
 & & Pick & Look & Clean & Heat & Cool & Pick2 & All & Score & Succ.\\
\midrule
\multicolumn{10}{l}{\textit{Closed-Source Model}} \\
Prompting& GPT-4o & 75.3 & 60.8 & 31.2 & 56.7 & 21.6 & 49.8 & 48.0& 31.8 & 23.7\\
Prompting& Gemini-2.5-Pro & 92.8 & 63.3 & 62.1 & 69.0 & 26.6 & 58.7 & 60.3& 42.5 & 35.9\\
\midrule
\multicolumn{10}{l}{\textit{Qwen2.5-1.5B-Instruct}} \\
Prompting& Qwen2.5 & 5.9 & 5.5 & 3.3 & 9.7 & 4.2 & 0.0 & 4.1 & 23.1 & 5.2\\
Prompting& ReAct & 17.4 & 20.5 & 15.7 & 6.2 & 7.7 & 2.0 & 12.8& 40.1& 11.3\\
Prompting& Reflexion & 35.3 & 22.2 & 21.7 & 13.6 & 19.4 & 3.7 & 21.8 & 55.8& 21.9\\
Outcome RL& PPO & 64.8\textsubscript{\textpm3.5} & 40.5\textsubscript{\textpm6.9} & 57.1\textsubscript{\textpm4.9} & 60.6\textsubscript{\textpm6.6} & 46.4\textsubscript{\textpm4.0} & 47.4\textsubscript{\textpm1.9} & 54.4\textsubscript{\textpm3.1}& 73.8\textsubscript{\textpm3.0} & 51.5\textsubscript{\textpm2.9} \\
Outcome RL& RLOO & 88.3\textsubscript{\textpm3.0} & 52.8\textsubscript{\textpm8.6} & 71.0\textsubscript{\textpm5.9} & 62.8\textsubscript{\textpm8.7} & 66.4\textsubscript{\textpm5.5} & 56.9\textsubscript{\textpm4.7} & 69.7\textsubscript{\textpm2.5}& 73.9\textsubscript{\textpm5.6}& 52.1\textsubscript{\textpm6.7}\\
Outcome RL& GRPO & 85.3\textsubscript{\textpm1.5} & 53.7\textsubscript{\textpm8.0} & 84.5\textsubscript{\textpm6.8} & 78.2\textsubscript{\textpm7.9} & 59.7\textsubscript{\textpm5.0} & 53.5\textsubscript{\textpm5.6} & 72.8\textsubscript{\textpm3.6}& 75.8\textsubscript{\textpm3.5} & 56.8\textsubscript{\textpm3.8}\\
Process RL& EMPG & 85.5\textsubscript{\textpm4.8} & 33.5\textsubscript{\textpm6.4} & 78.9\textsubscript{\textpm2.5} & 76.2\textsubscript{\textpm9.7} & 74.7\textsubscript{\textpm1.9} & 69.1\textsubscript{\textpm6.4} & 73.7\textsubscript{\textpm2.7} & 80.4\textsubscript{\textpm0.7} & 60.8\textsubscript{\textpm1.3}\\
Process RL& GiGPO & \underline{94.4}\textsubscript{\textpm5.9} & \underline{67.5}\textsubscript{\textpm4.6} & \textbf{94.8}\textsubscript{\textpm3.8} & \textbf{94.4}\textsubscript{\textpm7.8} & \underline{79.8}\textsubscript{\textpm4.7} & \underline{76.4}\textsubscript{\textpm5.4} & \underline{86.7}\textsubscript{\textpm1.7} & \underline{83.1}\textsubscript{\textpm1.6}& \underline{65.0}\textsubscript{\textpm3.2}\\

\rowcolor{gray!15}Process RL& \textbf{STAPO} & \textbf{97.7}\textsubscript{\textpm4.0} & \textbf{77.8}\textsubscript{\textpm4.8} & \underline{93.4}\textsubscript{\textpm0.5} & \underline{90.2}\textsubscript{\textpm9.4} & \textbf{85.6}\textsubscript{\textpm2.6} & \textbf{95.1}\textsubscript{\textpm0.4} & \textbf{92.2}\textsubscript{\textpm1.2} & \textbf{85.9}\textsubscript{\textpm0.6}& \textbf{69.0}\textsubscript{\textpm2.5}\\

\midrule
\multicolumn{10}{l}{\textit{Qwen2.5-7B-Instruct}} \\
Prompting& Qwen2.5 & 33.4 & 21.6 & 19.3 & 6.9 & 2.8 & 3.2 & 14.8 & 26.4 & 7.8\\
Prompting& ReAct & 48.5 & 35.4 & 34.3 & 13.2 & 18.2 & 17.6 & 31.2 & 46.2 & 19.5\\
Prompting& Reflexion & 62.0 & 41.6 & 44.9 & 30.9 & 36.3 & 23.8 & 42.7& 58.1& 28.8\\
Outcome RL& PPO & 92.3\textsubscript{\textpm4.0} & 64.0\textsubscript{\textpm8.4} & 92.5\textsubscript{\textpm2.4} & \underline{89.5}\textsubscript{\textpm7.0} & 80.3\textsubscript{\textpm2.0} & 68.8\textsubscript{\textpm8.3} & 80.4\textsubscript{\textpm2.7} & 81.4\textsubscript{\textpm3.1}& 68.7\textsubscript{\textpm5.1}\\
Outcome RL& RLOO & 87.6\textsubscript{\textpm4.3} & 78.2\textsubscript{\textpm8.3} & 87.3\textsubscript{\textpm5.8} & 81.3\textsubscript{\textpm7.6} & 71.9\textsubscript{\textpm5.2} & 48.9\textsubscript{\textpm8.4} & 75.5\textsubscript{\textpm4.6} & 80.3\textsubscript{\textpm3.2} & 65.7\textsubscript{\textpm4.0}\\
Outcome RL& GRPO & 90.8\textsubscript{\textpm5.1} & 66.1\textsubscript{\textpm6.7} & 89.3\textsubscript{\textpm5.4} & 74.7\textsubscript{\textpm6.9} & 72.5\textsubscript{\textpm5.4} & 64.7\textsubscript{\textpm7.3} & 77.6\textsubscript{\textpm5.2} & 79.3\textsubscript{\textpm2.8} & 66.1\textsubscript{\textpm3.7}\\
Process RL& EMPG & 92.9\textsubscript{\textpm2.9} & 75.2\textsubscript{\textpm3.8} & 74.8\textsubscript{\textpm3.9} & 86.3\textsubscript{\textpm4.7} & 73.7\textsubscript{\textpm2.6} & 65.3\textsubscript{\textpm5.8} & 78.5\textsubscript{\textpm1.7} & 81.0\textsubscript{\textpm1.4} & 69.3\textsubscript{\textpm0.5}\\
Process RL& GiGPO & \underline{97.7}\textsubscript{\textpm1.6} & \underline{82.7}\textsubscript{\textpm7.9} & \textbf{98.8}\textsubscript{\textpm1.6} & 83.7\textsubscript{\textpm7.2} & \underline{89.3}\textsubscript{\textpm8.2} & \underline{79.2}\textsubscript{\textpm6.6} & \underline{90.8}\textsubscript{\textpm1.3} & \underline{84.4}\textsubscript{\textpm2.9} & \underline{72.8}\textsubscript{\textpm3.2}\\

\rowcolor{gray!15}Process RL& \textbf{STAPO} & \textbf{97.8}\textsubscript{\textpm3.9} & \textbf{100.0}\textsubscript{\textpm0.0} & \textbf{98.8}\textsubscript{\textpm2.1} & \textbf{89.9}\textsubscript{\textpm7.5} & \textbf{92.7}\textsubscript{\textpm5.2} & \textbf{100.0}\textsubscript{\textpm0.0} & \textbf{96.9}\textsubscript{\textpm1.6} & \textbf{89.1}\textsubscript{\textpm1.3} & \textbf{77.6}\textsubscript{\textpm3.9}\\

\bottomrule
\end{tabular}
}
\caption{Performance on ALFWorld and WebShop. Results are averaged over 3 random seeds. For ALFWorld, we report the average success rate (\%) for each subtask as well as the overall result. For WebShop, we report both the average score and the average success rate (\%). The best and second-best results are \textbf{bold} and \underline{underlined}. Methods highlighted in gray are our implemented results. The remaining results are adopted from GiGPO~\citep{feng2025group} and EMPG~\citep{wang2025harnessing}.}
\label{tab:main1}
\end{table*}

\begin{table*}[t]
\centering
\small 
\resizebox{\textwidth}{!}{
\begin{tabular}{llcc@{\,\,\,}c|ccc@{\,\,\,}c|c}
\toprule
\multirow{2}{*}{Type} & \multirow{2}{*}{Method} & \multicolumn{3}{c|}{\textbf{Single-Hop QA}} & \multicolumn{4}{c|}{\textbf{Multi-Hop QA}} & \multirow{2}{*}{Avg.} \\
& & NQ$^{\dagger}$ & TriviaQA$^{\star}$ & PopQA$^{\star}$ & HotpotQA$^{\dagger}$ & 2Wiki$^{\star}$ & MuSiQue$^{\star}$ & Bamboogle$^{\star}$ & \\
\midrule
\multicolumn{9}{l}{\textit{Qwen2.5-3B-Instruct}} \\
Outcome RL& R1-Instruct  & 21.0 & 44.9 & 17.1 & 20.8 & 27.5 & 6.0 & 19.2 & 22.4 \\

Outcome RL&Search-R1   & 34.1 & 54.5 & 37.8 & 32.4 & 31.9 & 10.3 & 26.4 & 32.5 \\
Outcome RL&ZeroSearch  & 41.4 & 57.4 & \underline{44.8} & 27.4 & 30.0 & 9.8  & 11.1 & 31.7 \\
Process RL&StepSearch  & - & - & - & 34.5 & 32.0 & \textbf{17.4}  & 34.4 &  - \\
Process RL& GiGPO & \underline{42.0} & \underline{59.5} & 42.4 & \textbf{36.9} & \textbf{37.0} & 12.6 & \underline{64.1} & \underline{42.1} \\

\rowcolor{gray!15}Process RL& \textbf{STAPO} & \textbf{43.1} & \textbf{60.7} & \textbf{45.8} & \textbf{36.9} & \underline{36.4} & \underline{13.0} & \textbf{64.9} & \textbf{43.0} \\

\midrule
\multicolumn{9}{l}{\textit{Qwen2.5-7B-Instruct}} \\
Outcome RL&R1-Instruct  & 27.0 & 53.7 & 19.9 & 23.7 & 29.2 & 7.2 & 29.3 & 27.1 \\

Outcome RL&Search-R1   & 39.3 & 61.0 & 39.7 & 37.0 & 40.1 & 14.6 & 36.8 & 38.5 \\
Outcome RL&ZeroSearch  & 43.6 & 61.8 & \textbf{51.5} & 34.6 & 35.2 & 18.4 & 27.8 & 39.1 \\
Process RL&StepSearch  & - & - & - & 38.6 & 36.6 & \textbf{22.6}  & 40.0 &  - \\
Process RL& GiGPO & \underline{46.4} & \underline{64.7} & 46.1 & \underline{41.6}  & \underline{43.6} &  \underline{18.9} & \underline{68.9} & \underline{47.2} \\

\rowcolor{gray!15}Process RL& \textbf{STAPO} & \textbf{48.8} & \textbf{66.0} & \underline{48.7} & \textbf{43.0}  & \textbf{45.0} &  18.1 & \textbf{69.4} & \textbf{48.4} \\

\bottomrule
\end{tabular}
}
\caption{Performance on search-augmented QA tasks. STAPO is trained on NQ and HotpotQA. $\dagger$ and $\star$ indicate in-domain and out-of-domain datasets, respectively. The best and second-best results are \textbf{bold} and \underline{underlined}. Methods highlighted in gray are our implemented results. The remaining results are adopted from GiGPO~\citep{feng2025group}.}
\label{tab:main2}
\end{table*}

\subsection{Experiment Setup}
\paragraph{Benchmarks.} 
We first train the LLM agents on two challenging benchmarks: ALFWorld~\citep{shridhar2020alfworld} and WebShop~\citep{yao2022webshop}.
\textit{ALFWorld} is an embodied environment designed to assess the ability of LLM agents to perform multi-step decision-making. In each episode, the agent receives a text goal and must accomplish it through multi-turn interaction with the environment. It includes 3,827 task instances across six categories of common household activities: Pick \& Place (Pick), Examine in Light (Look), Clean \& Place (Clean), Heat \& Place (Heat), Cool \& Place (Cool), and Pick Two \& Place (Pick2).  
\textit{WebShop} is a complex, web-based interactive environment designed to test the LLM agents in realistic online shopping scenarios. To complete the task, the agent must interact with a simulated HTML-based shopping website to search for, navigate to, and ultimately purchase a suitable item. It contains over 1.1 million products and 12k user instructions, providing a rich and diverse action space.
In addition, we also evaluate the multi-turn tool calling performance of STAPO on \textit{search-augmented QA tasks}, including single-hop QA datasets (NQ~\citep{kwiatkowski2019natural}, TriviaQA~\citep{joshi2017triviaqa}, and PopQA~\citep{mallen2022not}) and multi-hop QA datasets (HotpotQA~\citep{yang2018hotpotqa}, 2Wiki~\citep{ho2020constructing}, MuSiQue~\citep{trivedi2022musique}, and Bamboogle~\citep{press2022measuring}).

\paragraph{Baselines.}

For ALFWorld and WebShop, we compare our approach with a range of competitive baselines under the same evaluation protocol: (1) Closed-source LLMs: GPT-4o~\citep{achiam2023gpt} and Gemini-2.5-Pro~\citep{team2023gemini}, which represent state-of-the-art capabilities in general-purpose reasoning and language understanding. (2) Prompting agents: ReAct~\citep{yao2023react} and Reflexion~\citep{shinn2024reflexion}, which rely on in-context prompting to guide multi-step behavior without parameter updates. (3) Outcome-supervised RL: Classical algorithms optimized solely on sparse outcome rewards, including PPO~\citep{schulman2017proximal}, RLOO~\citep{kool2019buy,ahmadian2024back}, and GRPO~\citep{shao2024deepseekmath}. (4) Process-supervised RL: Recent works utilizing process signals to assist RL training, including EMPG\citep{wang2025harnessing} and GiGPO~\citep{feng2025group}.
For search-augmented QA, we compare STAPO with R1-Instruct, Search-R1~\citep{jin2025search}, ZeroSearch~\citep{sun2025zerosearch}, and StepSearch~\citep{wang2025stepsearch} beyond GiGPO.

\paragraph{Implementation Details.}
We use the Qwen2.5 series and Llama3.1-8B-Instruct as our base models. The weighting coefficient $\omega$ is set to $1$ with no further tuning\footnote{Consistent with the setting in GiGPO.}.
For ALFWorld and WebShop, we employ exact matching (via hashmaps) for anchor state grouping, set the rollout group size $N$ to 8, and the max turns to 50 and 15, respectively. 
For search-augmented QA, we follow the same settings in Search-R1~\citep{jin2025search} and use E5~\citep{wang2022text} as the retriever. We adopt similarity-based anchor state grouping (longest matching subsequence with a threshold of 0.9) following GiGPO. The rollout group size $N$ is set to 5 and the max turn is set to 4.
All training is conducted on a single node with 8 A100 GPUs. Unless otherwise specified, the IQR coefficient $\lambda$ is set to $1.5$ (adopted from Tukey's fences, the standard statistical convention for defining IQR outliers), and the weighting coefficients $\alpha$, $\beta$, and $\gamma$ (Equations~\ref{eq:jbase}--\ref{eq:jsel}) are kept equal.
Full training settings and hyperparameter details are provided in Appendix~\ref{appendix:ed}.

\begin{table}[t]
\centering
\small 
\begin{tabular}{l c c}
\toprule
Variant & \textbf{ALFWorld} & \textbf{WebShop} \\
\midrule
\multicolumn{3}{l}{\textit{\textbf{Localization Strategy}}} \\
w/ Shannon Entropy & 89.6 & 67.2 \\
\midrule
\multicolumn{3}{l}{\textit{\textbf{Masking Strategy}}} \\
w/ None & 86.7 & 65.0 \\
w/ Goal Only & 89.0 & 66.6 \\
w/ History Only & 90.9 & 68.2 \\
\midrule
\multicolumn{3}{l}{\textit{\textbf{Penalty Strategy}}} \\
w/ Single Penalty & 88.3 & 65.9 \\
w/ Inc Penalty Coef & 84.6 & 67.0 \\
\midrule
\textbf{w/ STAPO} & \textbf{92.2} & \textbf{69.0} \\
\bottomrule
\end{tabular}
\caption{Ablation study on ALFWorld and WebShop. The results represent success rate (\%).}
\label{tab:ablation_study}
\end{table}

\subsection{Results}
\paragraph{Performance on ALFWorld and WebShop.}
Table~\ref{tab:main1} demonstrates the strong performance of STAPO on both ALFWorld and WebShop, where it consistently and significantly outperforms all baselines across different base model scales. Compared to the state-of-the-art method GiGPO, STAPO achieves superior results on ALFWorld, reaching near-perfect performance (96.9\%) with Qwen2.5-7B-Instruct. On WebShop, STAPO also shows a clear advantage, improving the success rate from 65.0\% (GiGPO) to 69.0\% with Qwen2.5-1.5B-Instruct. These results suggest that precise localization and optimization of outlier steps effectively enhance agent performance in long-horizon tasks. See Appendix~\ref{sec:GA} for further experiments on STAPO's generalization to other group-based RL methods (e.g., DAPO) and its adaptation across different model architectures.

\paragraph{Performance on Search-Augmented QA Tasks.} 
As shown in Table~\ref{tab:main2}, STAPO achieves consistent gains on multi-turn search-augmented QA tasks. While it outperforms strong baselines on average, the improvement over GiGPO ($\approx$ 1\%) is more modest than on ALFWorld and WebShop. We attribute this to the shorter interaction horizons of search-augmented QA tasks, which naturally mitigate trajectory neglect and thus limit the gains from STAPO’s selective optimization.
Table~\ref{tab:appendix_threshold_ablation} in Appendix~\ref{sec:appendix_robustness} reports robustness of QA performance to the similarity threshold used for anchor state grouping.

\begin{figure}[t]
\centering
  \includegraphics[width=\columnwidth]{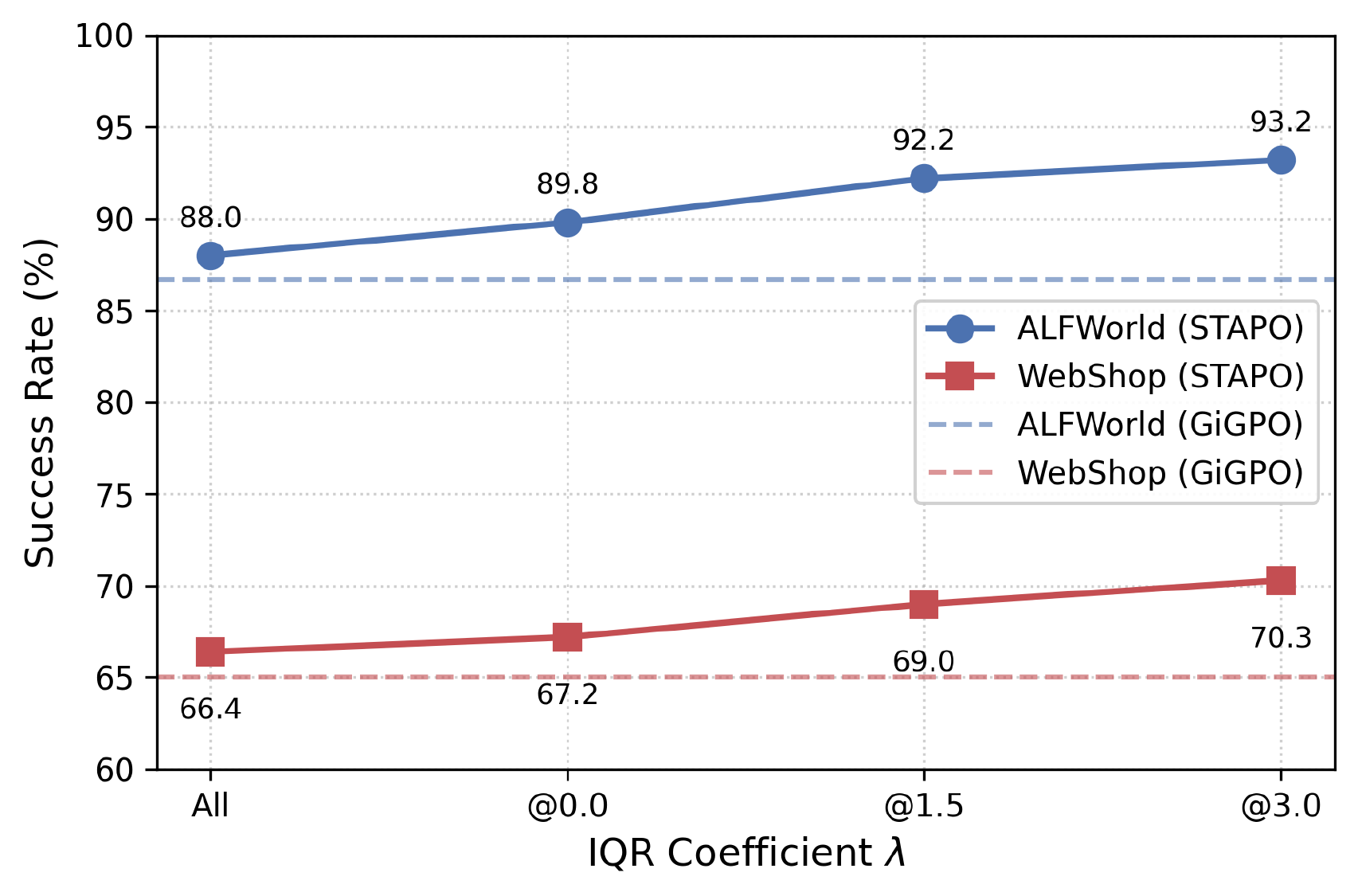}
  \caption{Ablation study of localization sensitivity. ``All'' means non-selective.}
  \label{fig:fig_as_iqr}
\end{figure}

\subsection{Ablation Study}
We validate the effectiveness of STAPO through ablation studies on localization, masking, and penalty strategies using Qwen2.5-1.5B-Instruct, with a sensitivity analysis of the weighting coefficients ($\alpha$ and $\gamma$) provided in Appendix~\ref{sec:appendix_SAW}.

\paragraph{Localization Strategy.} 
We evaluate the proposed outlier localization mechanism by examining the uncertainty metric choice and the sensitivity of the localization criterion.

\begin{itemize}
    \item \textbf{Entropy Type.} 
    As shown in the top block of Table~\ref{tab:ablation_study}, using normalized entropy for localization consistently outperforms raw Shannon entropy. This result indicates that normalized entropy more effectively decouples state complexity from agent uncertainty, enabling more accurate identification of true outlier steps, whereas raw Shannon entropy is more susceptible to state-dependent noise.
    
    \item \textbf{Localization Sensitivity.} 
    Figure~\ref{fig:fig_as_iqr} shows the effect of the IQR coefficient $\lambda$ controlling outlier strictness. STAPO consistently outperforms GiGPO across all settings, including ``All'', and performance improves as $\lambda$ increases from 0.0 to 3.0, suggesting that selectively optimizing a smaller set of confidently located outlier steps is beneficial.
\end{itemize}

\begin{figure}[t]
\centering
  \includegraphics[width=\columnwidth]{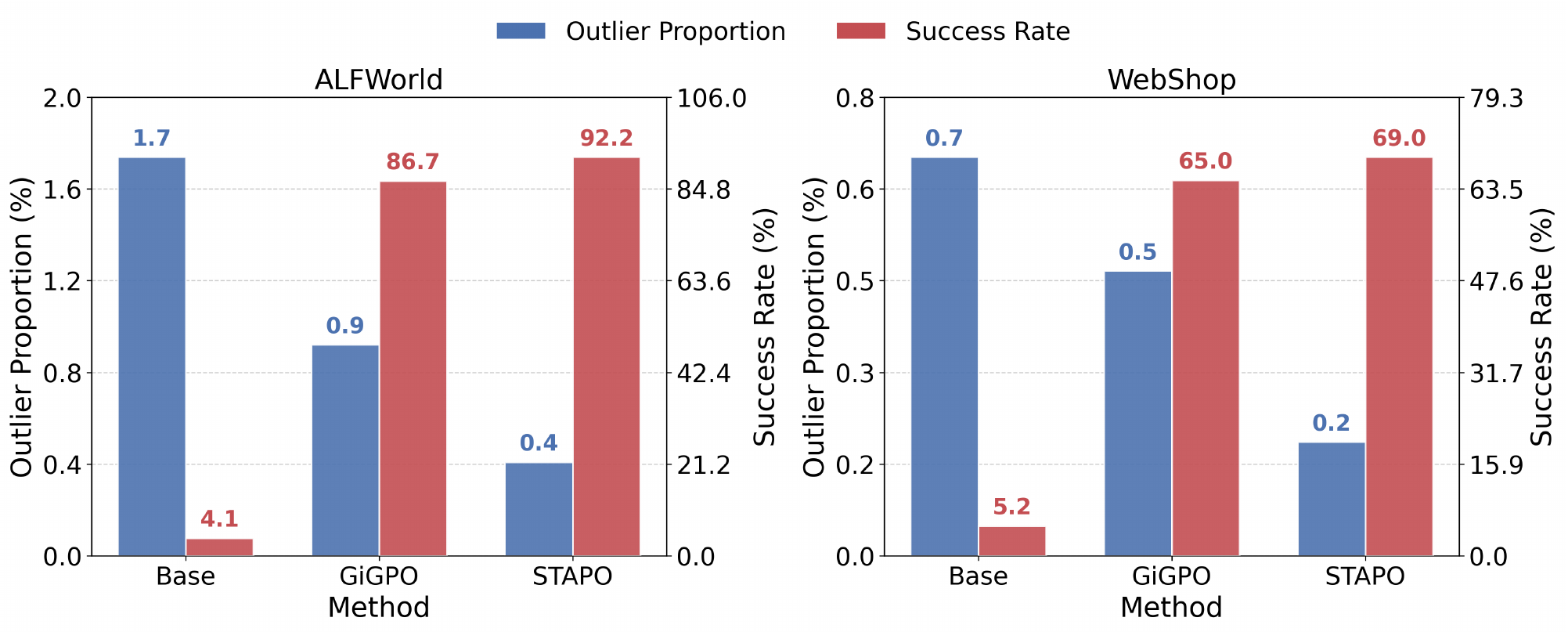}
  \caption{Attribution analysis on ALFWorld and WebShop. The blue bars (left axis) represent the proportion of outlier steps, while the red bars (right axis) indicate the success rate.}
  \label{fig:fig_aa}
\end{figure}

\begin{figure}[t]
\centering
  \includegraphics[width=\columnwidth]{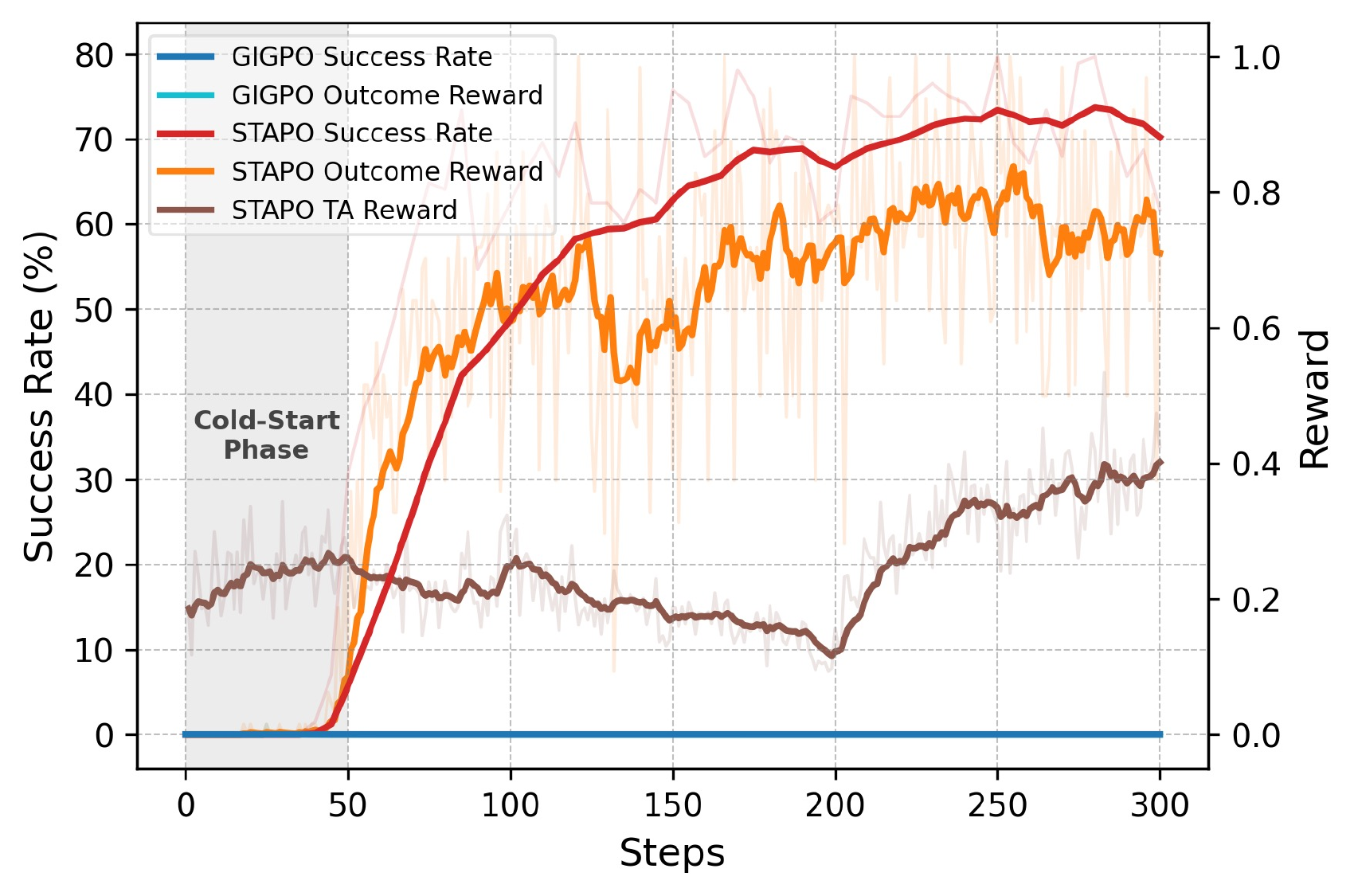}
  \caption{Training dynamics on WebShop using LLaMA-3.1-8B-Instruct. Solid lines denote exponential moving averages. “TA” refers to \textit{Trajectory-Aware}.}
  \label{fig:fig_eg}
\end{figure}

\paragraph{Masking Strategy.}
We compare different input masking strategies for reward calculation, as shown in the {middle block} of Table~\ref{tab:ablation_study}. Masking both the goal and interaction history (STAPO) consistently achieves the best performance. Notably, ``w/ History Only'' outperforms ``w/ Goal Only'', and both surpass the ``w/ None'' baseline (GiGPO), indicating that historical context is more critical than goal information alone for long-horizon decision-making. The further improvement from combining both signals highlights the importance of jointly reinforcing goal awareness and trajectory consistency to mitigate trajectory neglect.

\paragraph{Penalty Strategy.}
The {bottom block} of Table~\ref{tab:ablation_study} examines the effect of the trajectory-independent penalty. Compared to employing a standard global KL regularization (``w/ Single Reg''), simply increasing its weight (``w/ Inc Reg Coef'') can lead to performance degradation (e.g., 88.3\% $\to$ 84.6\% on ALFWorld), indicating that overly strong global constraints may hinder effective learning. In contrast, STAPO achieves the best overall results by applying the penalty selectively at outlier steps. This design encourages anchor policy updates to the reference policy at steps prone to trajectory neglect, while avoiding unnecessary constraints on well-behaved steps. As a result, the increase in $R_{TA}$ reliably enhanced trajectory focus rather than unintended effects such as mode collapse or overly conservative policy updates.

\subsection{Attribution Analysis}
To gain deeper insights into STAPO's mechanism, we statistically analyze the proportion of outlier steps generated by the Qwen2.5-1.5B-Instruct model on the test sets before and after training. As shown in Figure~\ref{fig:fig_aa}, the model's performance exhibits a negative correlation with the proportion of outliers. Compared to GiGPO, STAPO not only achieves better results but also yields trajectories with a significantly lower proportion of outliers. This result demonstrates that STAPO enhances performance fundamentally by mitigating low-quality actions caused by trajectory neglect. Refer to Appendix~\ref{sec:appendix_ICO} for an analysis of the internal composition (high vs. low entropy) of these outliers. Appendix~\ref{sec:appendix_causal} further reports test-time KL divergence between full and trajectory-blind prompts together with average trajectory attribution scores, as complementary evidence of increased reliance on trajectory.

\subsection{Exploration Guidance}
STAPO demonstrates superior exploration capability, particularly in cold-start scenarios. In challenging settings such as training Llama3.1-8B-Instruct on WebShop, the weak foundational policy fails to obtain any sparse outcome rewards, causing RL baselines like GiGPO to fail in optimizing effectively (as shown by the flatlined zero success rate in Figure~\ref{fig:fig_eg}). Conversely, STAPO's trajectory-aware reward $R_{TA}$ provides continuous supervisory signals. Especially in the initial phase (e.g., steps 0-50) where external outcome rewards are absent, $R_{TA}$ offers non-zero guidance, effectively steering the model towards meaningful actions and breaking the exploration deadlock.

\section{Conclusion}
In this work, we identify {trajectory neglect} as a primary cause of low-quality actions in long-horizon tasks and propose {normalized entropy} to precisely locate it by decoupling state complexity from agent confidence. Building on this, we introduce {STAPO}, a framework that selectively optimizes outliers via a joint mechanism of trajectory-aware reward and trajectory-independent penalty. Extensive experiments demonstrate that STAPO achieves superior performance and robustness across diverse benchmarks, providing a strong and competitive approach for trajectory-aware LLM agent training.

\section*{Limitations}
Regarding training efficiency, calculating the trajectory-aware reward and trajectory-independent penalty requires constructing masked inputs and performing additional forward passes. However, as detailed in Appendix~\ref{sec:appendix_TE}, this computational overhead is marginal ($\sim$5.7\%) and proves significantly more cost-effective than scaling the group size.

Our main experiments focus on text-based agents. Appendix~\ref{sec:appendix_vlm} provides additional pilot results on vision-language agents, but broader multimodal settings and outlier behavior under diverse visual interfaces remain to be fully characterized.

Normalized entropy assumes sufficient within-batch support for each anchor state: when a state is visited only once in a batch, intra-group normalization is ill-defined and the statistic effectively reduces to raw Shannon entropy, which again conflates state complexity with confidence. Environments with very large admissible action spaces make state revisitation rarer and can exacerbate this boundary behavior, occasionally yielding false positives or false negatives in outlier localization.

\section*{Ethical considerations}
In this work, we introduce a framework to enhance the performance of LLM agents in long-horizon tasks. While improving agent capabilities is crucial for beneficial applications such as digital assistants and automated web navigation, we acknowledge that more capable agents could potentially be exploited for malicious purposes, such as automated cyber-attacks or large-scale disinformation campaigns. However, our research is conducted strictly using established, public academic benchmarks (ALFWorld, WebShop, and QA datasets) that do not contain personally identifiable information (PII) or offensive content.

\section*{Acknowledgements}
This work was supported by the National Natural Science Foundation of China under Grant 42394060 and 42394064, Ant Group Research Fund, and Information Technology Center and State Key Lab of CAD\&CG, Zhejiang University.

\bibliography{acl2026_conference}

@book{sutton2018reinforcement,
  title={Reinforcement Learning: An Introduction},
  author={Sutton, Richard S and Barto, Andrew G},
  year={2018},
  publisher={MIT press}
}

@article{schulman2017proximal,
  title={Proximal policy optimization algorithms},
  author={Schulman, John and Wolski, Filip and Dhariwal, Prafulla and Radford, Alec and Klimov, Oleg},
  journal={arXiv preprint arXiv:1707.06347},
  year={2017}
}

@inproceedings{tan2024cradle,
  title={Cradle: Empowering foundation agents towards general computer control},
  author={Tan, Weihao and Zhang, Wentao and Xu, Xinrun and Xia, Haochong and Ding, Gang and Li, Boyu and Zhou, Bohan and Yue, Junpeng and Jiang, Jiechuan and Li, Yewen and others},
  booktitle={NeurIPS 2024 Workshop on Open-World Agents},
  year={2024}
}

@article{achiam2023gpt,
  title={{GPT-4} technical report},
  author={Achiam, Josh and Adler, Steven and Agarwal, Sandhini and Ahmad, Lama and Akkaya, Ilge and Aleman, Florencia Leoni and Almeida, Diogo and Altenschmidt, Janko and Altman, Sam and Anadkat, Shyamal and others},
  journal={arXiv preprint arXiv:2303.08774},
  year={2023}
}

@article{wang2024mobile,
  title={{Mobile-Agent-v2}: Mobile Device Operation Assistant with Effective Navigation via Multi-Agent Collaboration},
  author={Wang, Junyang and Xu, Haiyang and Jia, Haitao and Zhang, Xi and Yan, Ming and Shen, Weizhou and Zhang, Ji and Huang, Fei and Sang, Jitao},
  journal={Advances in Neural Information Processing Systems},
  volume={37},
  pages={2686--2710},
  year={2024}
}

@article{team2023gemini,
  title={Gemini: A family of highly capable multimodal models},
  author={Team, Gemini and Anil, Rohan and Borgeaud, Sebastian and Alayrac, Jean-Baptiste and Yu, Jiahui and Soricut, Radu and Schalkwyk, Johan and Dai, Andrew M and Hauth, Anja and Millican, Katie and others},
  journal={arXiv preprint arXiv:2312.11805},
  year={2023}
}

@article{zhang2024ufo,
  title={{UFO}: A {UI}-Focused Agent for Windows {OS} Interaction},
  author={Zhang, Chaoyun and Li, Liqun and He, Shilin and Zhang, Xu and Qiao, Bo and Qin, Si and Ma, Minghua and Kang, Yu and Lin, Qingwei and Rajmohan, Saravan and others},
  journal={arXiv preprint arXiv:2402.07939},
  year={2024}
}

@article{
wang2023voyager,
title={Voyager: An Open-Ended Embodied Agent with Large Language Models},
author={Guanzhi Wang and Yuqi Xie and Yunfan Jiang and Ajay Mandlekar and Chaowei Xiao and Yuke Zhu and Linxi Fan and Anima Anandkumar},
journal={Transactions on Machine Learning Research},
issn={2835-8856},
year={2024},
}

@inproceedings{
furuta2024multimodal,
title={Multimodal Web Navigation with Instruction-Finetuned Foundation Models},
author={Hiroki Furuta and Kuang-Huei Lee and Ofir Nachum and Yutaka Matsuo and Aleksandra Faust and Shixiang Shane Gu and Izzeddin Gur},
booktitle={The Twelfth International Conference on Learning Representations},
year={2024}
}

@inproceedings{brohan2023rt,
  title={{RT-2}: Vision-Language-Action Models Transfer Web Knowledge to Robotic Control},
  author={Zitkovich, Brianna and Yu, Tianhe and Xu, Sichun and Xu, Peng and Xiao, Ted and Xia, Fei and Wu, Jialin and Wohlhart, Paul and Welker, Stefan and Wahid, Ayzaan and others},
  booktitle={Conference on Robot Learning},
  pages={2165--2183},
  year={2023},
  organization={PMLR}
}

@inproceedings{
xie2024osworld,
title={{OSWorld}: Benchmarking Multimodal Agents for Open-Ended Tasks in Real Computer Environments},
author={Tianbao Xie and Danyang Zhang and Jixuan Chen and Xiaochuan Li and Siheng Zhao and Ruisheng Cao and Toh Jing Hua and Zhoujun Cheng and Dongchan Shin and Fangyu Lei and Yitao Liu and Yiheng Xu and Shuyan Zhou and Silvio Savarese and Caiming Xiong and Victor Zhong and Tao Yu},
booktitle={The Thirty-eight Conference on Neural Information Processing Systems Datasets and Benchmarks Track},
year={2024},
}

@article{wei2022chain,
  title={Chain-of-thought prompting elicits reasoning in large language models},
  author={Wei, Jason and Wang, Xuezhi and Schuurmans, Dale and Bosma, Maarten and Xia, Fei and Chi, Ed and Le, Quoc V and Zhou, Denny and others},
  journal={Advances in Neural Information Processing Systems},
  volume={35},
  pages={24824--24837},
  year={2022}
}

@inproceedings{zhang2023you,
  title={You Only Look at Screens: Multimodal Chain-of-Action Agents},
  author={Zhang, Zhuosheng and Zhang, Aston},
  booktitle={Findings of the Association for Computational Linguistics ACL 2024},
  pages={3132--3149},
  year={2024}
}

@inproceedings{
yao2023react,
title={{ReAct}: Synergizing Reasoning and Acting in Language Models},
author={Shunyu Yao and Jeffrey Zhao and Dian Yu and Nan Du and Izhak Shafran and Karthik R Narasimhan and Yuan Cao},
booktitle={The Eleventh International Conference on Learning Representations },
year={2023},
}

@article{schick2023toolformer,
  title={Toolformer: Language Models Can Teach Themselves to Use Tools},
  author={Schick, Timo and Dwivedi-Yu, Jane and Dess{\`\i}, Roberto and Raileanu, Roberta and Lomeli, Maria and Hambro, Eric and Zettlemoyer, Luke and Cancedda, Nicola and Scialom, Thomas},
  journal={Advances in Neural Information Processing Systems},
  volume={36},
  pages={68539--68551},
  year={2023}
}

@article{shinn2024reflexion,
  title={Reflexion: Language agents with verbal reinforcement learning},
  author={Shinn, Noah and Cassano, Federico and Gopinath, Ashwin and Narasimhan, Karthik and Yao, Shunyu},
  journal={Advances in Neural Information Processing Systems},
  volume={36},
  year={2024}
}

@article{yu2025dapo,
  title={{DAPO}: An open-source {LLM} reinforcement learning system at scale},
  author={Yu, Qiying and Zhang, Zheng and Zhu, Ruofei and Yuan, Yufeng and Zuo, Xiaochen and Yue, Yu and Fan, Tiantian and Liu, Gaohong and Liu, Lingjun and Liu, Xin and others},
  journal={arXiv preprint arXiv:2503.14476},
  year={2025}
}

@article{li2024embodied,
  title={Embodied Agent Interface: Benchmarking {LLMs} for Embodied Decision Making},
  author={Li, Manling and Zhao, Shiyu and Wang, Qineng and Wang, Kangrui and Zhou, Yu and Srivastava, Sanjana and Gokmen, Cem and Lee, Tony and Li, Erran Li and Zhang, Ruohan and others},
  journal={Advances in Neural Information Processing Systems},
  volume={37},
  pages={100428--100534},
  year={2024}
}

@article{yao2022webshop,
  title={{WebShop}: Towards Scalable Real-World Web Interaction with Grounded Language Agents},
  author={Yao, Shunyu and Chen, Howard and Yang, John and Narasimhan, Karthik},
  journal={Advances in Neural Information Processing Systems},
  volume={35},
  pages={20744--20757},
  year={2022}
}

@inproceedings{ahmadian2024back,
  title={Back to Basics: Revisiting REINFORCE Style Optimization for Learning from Human Feedback in {LLMs}},
  author={Ahmadian, Arash and Cremer, Chris and Gall{\'e}, Matthias and Fadaee, Marzieh and Kreutzer, Julia and Pietquin, Olivier and {\"U}st{\"u}n, Ahmet and Hooker, Sara},
  booktitle={Proceedings of the 62nd Annual Meeting of the Association for Computational Linguistics (Volume 1: Long Papers)},
  year={2024},
  pages={12248--12267},
}

@inproceedings{kool2019buy,
  title={Buy 4 reinforce samples, get a baseline for free!},
  author={Kool, Wouter and van Hoof, Herke and Welling, Max},
  booktitle={ICLR 2019 Workshop},
  year={2019},
}

@article{shao2024deepseekmath,
  title={{DeepSeekMath}: Pushing the Limits of Mathematical Reasoning in Open Language Models},
  author={Shao, Zhihong and Wang, Peiyi and Zhu, Qihao and Xu, Runxin and Song, Junxiao and Bi, Xiao and Zhang, Haowei and Zhang, Mingchuan and Li, YK and Wu, Y and others},
  journal={arXiv preprint arXiv:2402.03300},
  year={2024}
}

@article{jin2025search,
  title={{Search-R1}: Training {LLMs} to Reason and Leverage Search Engines with Reinforcement Learning},
  author={Jin, Bowen and Zeng, Hansi and Yue, Zhenrui and Wang, Dong and Zamani, Hamed and Han, Jiawei},
  journal={arXiv preprint arXiv:2503.09516},
  year={2025}
}

@inproceedings{
shridhar2020alfworld,
title={{ALFWorld}: Aligning Text and Embodied Environments for Interactive Learning},
author={Mohit Shridhar and Xingdi Yuan and Marc-Alexandre Cote and Yonatan Bisk and Adam Trischler and Matthew Hausknecht},
booktitle={International Conference on Learning Representations},
year={2021},
}

@article{sun2025zerosearch,
  title={{ZeroSearch}: Incentivize the Search Capability of LLMs without Searching},
  author={Sun, Hao and Qiao, Zile and Guo, Jiayan and Fan, Xuanbo and Hou, Yingyan and Jiang, Yong and Xie, Pengjun and Huang, Fei and Zhang, Yan},
  journal={arXiv preprint arXiv:2505.04588},
  year={2025}
}

@inproceedings{
gou2025navigating,
title={Navigating the Digital World as Humans Do: Universal Visual Grounding for {GUI} Agents},
author={Boyu Gou and Ruohan Wang and Boyuan Zheng and Yanan Xie and Cheng Chang and Yiheng Shu and Huan Sun and Yu Su},
booktitle={The Thirteenth International Conference on Learning Representations},
year={2025}
}

@article{zheng2024gpt,
  title={{GPT-4V} (ision) is a generalist web agent, if grounded},
  author={Zheng, Boyuan and Gou, Boyu and Kil, Jihyung and Sun, Huan and Su, Yu},
  journal={arXiv preprint arXiv:2401.01614},
  year={2024}
}

@article{kwiatkowski2019natural,
  title={Natural questions: a benchmark for question answering research},
  author={Kwiatkowski, Tom and Palomaki, Jennimaria and Redfield, Olivia and Collins, Michael and Parikh, Ankur and Alberti, Chris and Epstein, Danielle and Polosukhin, Illia and Devlin, Jacob and Lee, Kenton and others},
  journal={Transactions of the Association for Computational Linguistics},
  volume={7},
  pages={453--466},
  year={2019},
  publisher={MIT Press One Rogers Street, Cambridge, MA 02142-1209, USA journals-info~…}
}

@article{yang2018hotpotqa,
  title={{HotpotQA}: A dataset for diverse, explainable multi-hop question answering},
  author={Yang, Zhilin and Qi, Peng and Zhang, Saizheng and Bengio, Yoshua and Cohen, William W and Salakhutdinov, Ruslan and Manning, Christopher D},
  journal={arXiv preprint arXiv:1809.09600},
  year={2018}
}

@article{trivedi2022musique,
  title={{MuSiQue}: Multihop Questions via Single-hop Question Composition},
  author={Trivedi, Harsh and Balasubramanian, Niranjan and Khot, Tushar and Sabharwal, Ashish},
  journal={Transactions of the Association for Computational Linguistics},
  volume={10},
  pages={539--554},
  year={2022},
  publisher={MIT Press One Broadway, 12th Floor, Cambridge, Massachusetts 02142, USA~…}
}

@article{press2022measuring,
  title={Measuring and narrowing the compositionality gap in language models},
  author={Press, Ofir and Zhang, Muru and Min, Sewon and Schmidt, Ludwig and Smith, Noah A and Lewis, Mike},
  journal={arXiv preprint arXiv:2210.03350},
  year={2022}
}

@article{mallen2022not,
  title={When not to trust language models: Investigating effectiveness of parametric and non-parametric memories},
  author={Mallen, Alex and Asai, Akari and Zhong, Victor and Das, Rajarshi and Khashabi, Daniel and Hajishirzi, Hannaneh},
  journal={arXiv preprint arXiv:2212.10511},
  year={2022}
}

@article{ho2020constructing,
  title={Constructing a multi-hop qa dataset for comprehensive evaluation of reasoning steps},
  author={Ho, Xanh and Nguyen, Anh-Khoa Duong and Sugawara, Saku and Aizawa, Akiko},
  journal={arXiv preprint arXiv:2011.01060},
  year={2020}
}

@article{joshi2017triviaqa,
  title={{TriviaQA}: A large scale distantly supervised challenge dataset for reading comprehension},
  author={Joshi, Mandar and Choi, Eunsol and Weld, Daniel S and Zettlemoyer, Luke},
  journal={arXiv preprint arXiv:1705.03551},
  year={2017}
}

@article{wang2025stepsearch,
  title={{StepSearch}: Igniting {LLMs} Search Ability via Step-Wise Proximal Policy Optimization},
  author={Wang, Ziliang and Zheng, Xuhui and An, Kang and Ouyang, Cijun and Cai, Jialu and Wang, Yuhang and Wu, Yichao},
  journal={arXiv preprint arXiv:2505.15107},
  year={2025}
}

@article{wang2022text,
  title={Text embeddings by weakly-supervised contrastive pre-training},
  author={Wang, Liang and Yang, Nan and Huang, Xiaolong and Jiao, Binxing and Yang, Linjun and Jiang, Daxin and Majumder, Rangan and Wei, Furu},
  journal={arXiv preprint arXiv:2212.03533},
  year={2022}
}

@article{wang2025harnessing,
  title={Harnessing uncertainty: Entropy-modulated policy gradients for long-horizon llm agents},
  author={Wang, Jiawei and Liu, Jiacai and Fu, Yuqian and Li, Yingru and Wang, Xintao and Lin, Yuan and Yue, Yu and Zhang, Lin and Wang, Yang and Wang, Ke},
  journal={arXiv preprint arXiv:2509.09265},
  year={2025}
}

@article{zhang2025rlvmr,
  title={Rlvmr: Reinforcement learning with verifiable meta-reasoning rewards for robust long-horizon agents},
  author={Zhang, Zijing and Chen, Ziyang and Li, Mingxiao and Tu, Zhaopeng and Li, Xiaolong},
  journal={arXiv preprint arXiv:2507.22844},
  year={2025}
}

@article{xu2025epo,
  title={EPO: Entropy-regularized Policy Optimization for LLM Agents Reinforcement Learning},
  author={Xu, Wujiang and Zhao, Wentian and Wang, Zhenting and Li, Yu-Jhe and Jin, Can and Jin, Mingyu and Mei, Kai and Wan, Kun and Metaxas, Dimitris N},
  journal={arXiv preprint arXiv:2509.22576},
  year={2025}
}

@article{agarwal2025unreasonable,
  title={The unreasonable effectiveness of entropy minimization in llm reasoning},
  author={Agarwal, Shivam and Zhang, Zimin and Yuan, Lifan and Han, Jiawei and Peng, Hao},
  journal={arXiv preprint arXiv:2505.15134},
  year={2025}
}

@article{cui2025entropy,
  title={The entropy mechanism of reinforcement learning for reasoning language models},
  author={Cui, Ganqu and Zhang, Yuchen and Chen, Jiacheng and Yuan, Lifan and Wang, Zhi and Zuo, Yuxin and Li, Haozhan and Fan, Yuchen and Chen, Huayu and Chen, Weize and others},
  journal={arXiv preprint arXiv:2505.22617},
  year={2025}
}

@article{feng2025group,
  title={Group-in-group policy optimization for llm agent training},
  author={Feng, Lang and Xue, Zhenghai and Liu, Tingcong and An, Bo},
  journal={arXiv preprint arXiv:2505.10978},
  year={2025}
}

@article{kokhlikyan2020captum,
  title={Captum: A unified and generic model interpretability library for PyTorch. arXiv},
  author={Kokhlikyan, Narine and Miglani, Vivek and Martin, Miguel and Wang, Edward and Alsallakh, Bilal and Reynolds, Jonathan and Melnikov, Alexander and Kliushkina, Natalia and Araya, Carlos and Yan, Siqi and others},
  journal={arXiv preprint arXiv:2009.07896},
  volume={2},
  pages={5},
  year={2020}
}

@article{ji2025tree,
  title={Tree search for llm agent reinforcement learning},
  author={Ji, Yuxiang and Ma, Ziyu and Wang, Yong and Chen, Guanhua and Chu, Xiangxiang and Wu, Liaoni},
  journal={arXiv preprint arXiv:2509.21240},
  year={2025}
}
\clearpage
\appendix

\section{Case Study of Trajectory Neglect in Multi-Turn Interactions}
\label{sec:CSTN}

To motivate the necessity of trajectory-aware optimization, we analyze a failed trajectory of a Qwen2.5-1.5B-Instruct agent trained with GiGPO from ALFWorld. The task goal is \textit{``look at cd under the desklamp.''} Throughout this episode, the agent exhibits two primary failure modes correlated with trajectory neglect: (1) {Task Goal Neglect}, where the agent wanders to irrelevant locations (e.g., the bed) despite the goal explicitly constraining the context to a desklamp, and (2) {Interaction History Neglect}, resulting in cyclic and redundant actions.

\paragraph{Case: Neglect of Task Goal.}
As shown in Figure~\ref{case:goal_neglect}, during Steps 21--25, the agent successfully acquires cd 1. However, instead of attempting to place it under a desklamp, the agent begins examining the object and wanders to bed 1 (Step 23, highlighted with an underline), a location irrelevant to the desklamp. This indicates a degradation of the goal signal over long horizons.

\paragraph{Case: Neglect of Interaction History.}
As shown in Figure~\ref{case:history_neglect}, during Steps 34--38, the agent demonstrates a severe inability to track its recent actions. After identifying cd 2 on shelf 1, the agent performs the action \texttt{look at cd 2}. In the very next step, despite the observation remaining static, the agent repeats the exact same action (Step 35, highlighted with an underline). Furthermore, the agent later returns to desk 2, a location it had already visited in Step 25. This suggests that the agent is not effectively tracking its interaction history.

\section{Case Study: Normalized vs. Shannon Entropy}
\label{sec:appendix_CSNAE}
To empirically validate the effectiveness of normalized entropy in decoupling state complexity from model confidence, we present qualitative examples from the ALFWorld evaluation using the Qwen2.5-1.5B-Instruct base model. These cases demonstrate how normalized entropy avoids false positives in states with a large action space and reduces false negatives in states with a small action space, thereby providing a more reliable signal for outlier localization.

\paragraph{Case: Avoiding False Positives in States with a Large Action Space.}
As shown in Figure~\ref{case:avoiding_false}, at Step 27, the agent is positioned in a kitchen environment with a vast action space (32 selectable actions). The agent chooses to navigate to ``countertop 1'', a valid exploration step denoted by a positive advantage. However, due to the sheer number of admissible actions, the raw probability distribution is naturally dispersed, resulting in a high {Shannon entropy}. A standard threshold-based method (e.g., IQR) would incorrectly flag this valid step as an outlier (a ``false positive''). In contrast, {normalized entropy} calibrates this value against the group statistics of the same state. It correctly recognizes that high entropy is intrinsic to this specific state and classifies the step as an inlier, preventing the unnecessary penalization of valid exploration.

\paragraph{Case: Reducing False Negatives in States with a Small Action Space.}
As shown in Figure~\ref{case:reducing_false}, at Step 43, the agent operates in a restricted context with a limited action space (16 selectable actions). Here, the agent exhibits suboptimal behavior by closing an empty cabinet it has already inspected (interaction history neglect), a redundant action indicated by a negative advantage. Despite this low-quality decision, the limited number of choices results in a naturally sharper probability distribution and a relatively low {Shannon entropy}. Consequently, a standard threshold-based method (e.g., IQR) would likely miss this error (a ``false negative''). However, {normalized entropy} detects that the agent's raw Shannon entropy—while numerically low—is significantly higher than the baseline expected for this specific state. This anomaly correctly flags the step as an outlier, triggering the STAPO to optimize the behavior.

\section{Pseudocode for Outlier Localization}
\label{sec:appendix_pseudocode}
Algorithm~\ref{alg:outlier_localization} summarizes anchor state formation, normalized entropy calculation, and outlier mask generation.

\begin{algorithm}[t]
\caption{Outlier localization}
\label{alg:outlier_localization}
\begin{algorithmic}[1]
\Require Batch of steps with prompts $\{c_t^{(i)}\}$ and actions $\{a_t^{(i)}\}$, grouping rule $\textsc{AnchorKey}(\cdot)$, stability $\epsilon$, Tukey coefficient $\lambda$
\State Initialize empty groups $\{\mathcal{G}^S(\tilde{s})\}$
\For{each step $(i,t)$ in the batch}
    \State $\tilde{s} \gets \textsc{AnchorKey}(c_t^{(i)})$
    \State Append $(a_t^{(i)}, R_t^{(i)}, c_t^{(i)})$ to $\mathcal{G}^S(\tilde{s})$
\EndFor
\For{each group $\mathcal{G}^S(\tilde{s})$}
    \For{each $(a_t^{(i)}, R_t^{(i)}, c_t^{(i)}) \in \mathcal{G}^S(\tilde{s})$}
        \State Compute $H(a_t^{(i)})$ using Equation~\ref{eq:AE}
    \EndFor
    \State Compute $\mu_{\tilde{s}}, \sigma_{\tilde{s}}$ over $\{H(a_t^{(j)})\}$ in $\mathcal{G}^S(\tilde{s})$
    \For{each $(a_t^{(i)}, \dots) \in \mathcal{G}^S(\tilde{s})$}
        \State $H_n(a_t^{(i)}) \gets (H(a_t^{(i)}) - \mu_{\tilde{s}}) / (\sigma_{\tilde{s}} + \epsilon)$ \Comment{Equation~\ref{eq:RE}}
    \EndFor
\EndFor
\State Let $\mathcal{H}_n$ be the multiset of all $H_n$ values in the batch
\State Compute quartiles $Q_1, Q_3$ of $\mathcal{H}_n$; $\text{IQR} \gets Q_3 - Q_1$
\For{each step $(i,t)$}
    \State $M_t(a_t^{(i)}) \gets \mathbb{I}\bigl(H_n \notin [Q_1 - \lambda\cdot\text{IQR},\, Q_3 + \lambda\cdot\text{IQR}]\bigr)$ \Comment{Equation~\ref{eq:IM}}
\EndFor
\end{algorithmic}
\end{algorithm}

\section{Details for Empirical Analysis}
\label{sec:appendix_DEA}

In this paper, we adopt Feature Ablation (implemented via \texttt{FeatureAblation} in Captum~\citep{kokhlikyan2020captum}) as our method for calculating trajectory attribution scores for LLMs.
For ALFWorld and WebShop, we randomly select 128 samples from the test set and conduct empirical analysis on the trajectories generated by 8 rollouts per sample.

\section{Experiment Details}
\label{appendix:ed}
\subsection{Details of Training}
\label{appendix:td}
\paragraph{Hyperparameters for ALFWorld.}
The maximum prompt length is 2048 tokens, and the maximum response length is 512 tokens. Each episode allows up to 50 environment steps. The learning rate is set to 1e-6 for the actor. We adopt a rule-based reward, assigning a reward of 1 for success and 0 for failure. To handle invalid actions generated by the agent, we apply a reward penalty of -0.1. We use a group size of 8 and sample 16 different groups per rollout, resulting in a total of $16 \times 8 = 128$ environments. The rollout temperature is set to 1.0, while the validation temperature is set to 0.4. The mini-batch size is 256, and the weighting coefficient $\omega$ is fixed at $1$ without further tuning. The IQR coefficient $\lambda$ is set to $1.5$, and the weighting coefficients $\alpha$, $\beta$, and $\gamma$ (Equations~\ref{eq:jbase}--\ref{eq:jsel}) are all set to $0.01$.

\paragraph{Hyperparameters for WebShop.}
The maximum prompt length is 4096 tokens, and the maximum response length is 512 tokens. Each episode is limited to 15 environment steps. The learning rate is 1e-6 for the actor. We adopt a rule-based reward, assigning a reward of 1 for success and 0 for failure. Invalid actions are penalized with a reward of -0.1. We use a group size of 8 and sample 8 groups per rollout, totaling $8 \times 8 = 64$ environments. The rollout temperature is set to 1.0, while the validation temperature is set to 0.4. The mini-batch size is 64, and the weighting coefficient $\omega$ is fixed at $1$ without further tuning. The IQR coefficient $\lambda$ is set to $1.5$, and the weighting coefficients $\alpha$, $\beta$, and $\gamma$ (Equations~\ref{eq:jbase}--\ref{eq:jsel}) are all set to $0.01$.

\paragraph{Hyperparameters for Search-Augmented QA.}
The maximum prompt length is 4096 tokens, and the maximum response length is 512 tokens. The max turn is set to 4. The learning rate is 1e-6 for the actor. We adopt a rule-based reward, assigning a reward of 1 for success and 0 for failure. Invalid actions are penalized with a reward of -0.01. We set the train data size to 256 and use a group size of 5. Rollout and validation temperatures are set to 1.0 and 0.0, respectively. The mini-batch size is 512, and the weighting coefficient $\omega$ is set to $1$ without additional tuning. The IQR coefficient $\lambda$ is set to $1.5$, and the weighting coefficients $\alpha$, $\beta$, and $\gamma$ (Equations~\ref{eq:jbase}--\ref{eq:jsel}) are all set to $0.001$.

\paragraph{Computing Details.}
Qwen2.5-1.5B-Instruct uses 2×A100 GPUs and Qwen2.5-7B-Instruct uses 4×A100 GPUs, with training extending to 150 iterations on ALFWorld and 300 on WebShop. For search-augmented QA, Qwen2.5-3B-Instruct uses 4×A100 GPUs and Qwen2.5-7B-Instruct uses 8×A100 GPUs, each for 200 iterations.

\begin{table*}[t]
\centering
\small 
\resizebox{\textwidth}{!}{
\begin{tabular}{llccccccc|cc}
\toprule

\multirow{2}{*}{Type} & \multirow{2}{*}{Method} & \multicolumn{7}{c|}{\textbf{ALFWorld}} & \multicolumn{2}{c}{\textbf{WebShop}} \\
 & & Pick & Look & Clean & Heat & Cool & Pick2 & All & Score & Succ.\\
\midrule
\multicolumn{10}{l}{\textit{Qwen2.5-1.5B-Instruct}} \\
Outcome RL& DAPO & 88.1 & 61.4 & 82.5 & 90.1 & 83.9 & 69.5 & 80.8 & 84.6 & 66.1 \\
Process RL& GiGPO\textsubscript{dynamic} & \underline{93.3} & \textbf{75.0} & \underline{96.9} & \underline{93.8} & \underline{91.7} & \underline{81.8} & \underline{91.4} & \underline{87.5} & \underline{75.0} \\
Process RL& \textbf{STAPO\textsubscript{dynamic}} & \textbf{99.1} & \underline{72.2} & \textbf{100.0} & \textbf{100.0} & \textbf{95.8} & \textbf{93.5} & \textbf{96.1} & \textbf{91.3} & \textbf{80.2} \\
\bottomrule
\end{tabular}
}
\caption{Generalization performance on ALFWorld and WebShop when integrating strategies adapted from DAPO. The best and second-best results are \textbf{bold} and \underline{underlined}.}
\label{tab:GOGBRM}
\end{table*}

\begin{table*}[t]
\centering
\small 

\begin{tabular}{lccccccc}
\toprule

\multirow{2}{*}{Method} & \multicolumn{7}{c}{\textbf{ALFWorld}} \\
 & Pick & Look & Clean & Heat & Cool & Pick2 & All \\
\midrule
\multicolumn{8}{l}{\textit{Llama3.1-8B-Instruct}} \\
GiGPO & \underline{89.7} & \underline{83.3} & \underline{96.8} & \textbf{100.0} & \underline{82.6} & \underline{90.0} & \underline{91.4} \\

\textbf{STAPO} & \textbf{93.1} & \textbf{100.0} & \textbf{100.0} & \underline{89.5} & \textbf{95.7} & \textbf{95.0} & \textbf{95.3} \\

\midrule
\multicolumn{8}{l}{\textit{Qwen2.5-14B-Instruct}} \\
GiGPO & \textbf{99.1} & \underline{77.8} & \textbf{100.0} & \underline{89.9} & \underline{86.4} & \underline{90.1} & \underline{93.2} \\

\textbf{STAPO} & \textbf{99.1} & \textbf{91.7} & \textbf{100.0} & \textbf{100.0} & \textbf{94.4} & \textbf{91.7} & \textbf{97.2} \\

\bottomrule
\end{tabular}

\caption{Adaptation performance on ALFWorld across different model architectures and scales. The best and second-best results are \textbf{bold} and \underline{underlined}.}
\label{tab:ADMAS}
\end{table*}

\subsection{Prompts}
The prompts we use for STAPO are presented in Figures~\ref{prompt:alfworld_complete}--\ref{prompt:search_blind}. These prompt templates are constructed using Python-style string formatting, where placeholders enclosed in curly braces (\{\}) represent semantic slots. These placeholders, such as \{task\_description\}, \{step\_count\}, and \{current\_observation\}, are dynamically populated at runtime via Python's \texttt{.format()} function.

The <think> </think> block instructs the agent to explicitly perform step-by-step reasoning, thereby promoting chain-of-thought~\citep{wei2022chain} style deliberation. The <action> </action> block is used to clearly indicate the final action decision.
The search agent outputs reasoning traces within <think> </think>, issues search queries within <search> </search>, provides anwsers within <anwser> </anwser>. Retrieved evidence from the retriever is presented in <information> </information> tags.

\section{Generalization and Adaptation}
\label{sec:GA}

To further verify the versatility and robustness of STAPO, we conduct extensive experiments focusing on its compatibility with advanced RL methods and its scalability across different model architectures. The results are summarized in Table~\ref{tab:GOGBRM} and Table~\ref{tab:ADMAS}.

\paragraph{Generalization to Other Group-Based RL Methods.}
The STAPO framework is orthogonal to advancements in single-turn group-based RL, allowing it to seamlessly incorporate complementary techniques. To demonstrate this, we follow the setup of DAPO~\cite{yu2025dapo}, which enhances GRPO with dynamic sampling and clip-higher mechanisms. We integrate these strategies into both GiGPO and STAPO, yielding variants denoted as GiGPO\textsubscript{dynamic} and STAPO\textsubscript{dynamic}.
As shown in Table~\ref{tab:GOGBRM}, while GiGPO\textsubscript{dynamic} improves over the DAPO baseline, STAPO\textsubscript{dynamic} achieves significantly superior results. Specifically, on WebShop, STAPO\textsubscript{dynamic} attains a success rate of 80.2\%, outperforming GiGPO\textsubscript{dynamic} (75.0\%) by a substantial margin. On ALFWorld, it reaches a remarkable 96.1\% success rate. These results confirm that STAPO can effectively synergize with and amplify the benefits of DAPO, establishing a new state-of-the-art performance.

\paragraph{Adaptation across Different Model Architectures and Scales.}
We further evaluate the scalability of STAPO by applying it to models with larger scales and different architectures: Llama3.1-8B-Instruct and Qwen2.5-14B-Instruct. As presented in Table~\ref{tab:ADMAS}, STAPO consistently outperforms the GiGPO baseline. For Llama3.1-8B-Instruct, STAPO improves the overall success rate on ALFWorld from 91.4\% to 95.3\%. Similarly, on the larger Qwen2.5-14B-Instruct model, STAPO pushes the performance to a near-perfect 97.2\%, compared to 93.2\% for GiGPO. This consistent improvement highlights that the trajectory-aware optimization in STAPO is model-agnostic and remains highly effective as model scale increases.

\section{Robustness to QA Similarity Threshold}
\label{sec:appendix_robustness}
For search-augmented QA we use Qwen2.5-3B-Instruct and vary the longest matching subsequence similarity threshold for anchor state grouping from $0.9$ to $0.5$. Table~\ref{tab:appendix_threshold_ablation} shows per-dataset performance, the average over seven datasets, and the average group size. Performance stays relatively flat while group size grows, suggesting robustness to moderate grouping noise.

\begin{table*}[t]
\centering
\small
\resizebox{\textwidth}{!}{
\begin{tabular}{lcccccccccc}
\toprule
Thres. & NQ & TriviaQA & PopQA & HotpotQA & 2Wiki & MuSiQue & Bamboogle & Avg.\ Perf. & Avg.\ Group Size \\
\midrule
\multicolumn{10}{l}{\textit{Qwen2.5-3B-Instruct}} \\
0.9 & 43.1 & 60.7 & 45.8 & 36.9 & 36.4 & 13.0 & 64.9 & 43.0 & 2.21 \\
0.8 & 43.2 & 60.1 & 44.4 & 37.3 & 37.4 & 13.0 & 65.3 & 43.0 & 2.20 \\
0.7 & 45.2 & 60.6 & 45.8 & 37.5 & 33.5 & 12.8 & 62.9 & 42.6 & 2.37 \\
0.6 & 44.0 & 60.0 & 46.2 & 36.8 & 36.3 & 12.1 & 61.7 & 42.4 & 3.04 \\
0.5 & 43.8 & 61.1 & 46.3 & 38.1 & 36.8 & 13.2 & 60.9 & 42.9 & 3.15 \\
\bottomrule
\end{tabular}
}
\caption{Effect of anchor state similarity threshold on search-augmented QA tasks.}
\label{tab:appendix_threshold_ablation}
\end{table*}

\section{Sensitivity Analysis of Weighting Coefficients $\alpha$ and $\gamma$}
\label{sec:appendix_SAW}

To investigate the impact of the selective optimization components on model performance, we conduct a sensitivity analysis on the weighting coefficients for the trajectory-aware reward $\alpha$ and the trajectory-independent penalty $\gamma$.

\paragraph{Settings.}
We conduct this analysis on the WebShop benchmark using the Qwen2.5-1.5B-Instruct model. Based on the hypothesis that the reward signal and the penalty constraint should be balanced to prevent optimization instability, we impose the constraint $\alpha = \gamma$ and sweep their values across the set $\{0, 0.005, 0.01, 0.02, 0.04\}$. The coefficient for the standard global KL penalty, $\beta$, is fixed at 0.01 throughout the experiments. Note that the setting $\alpha=\gamma=0$ corresponds to the GiGPO baseline, where no selective trajectory-aware optimization is applied.

\begin{figure}[t]
\centering
  \includegraphics[width=\columnwidth]{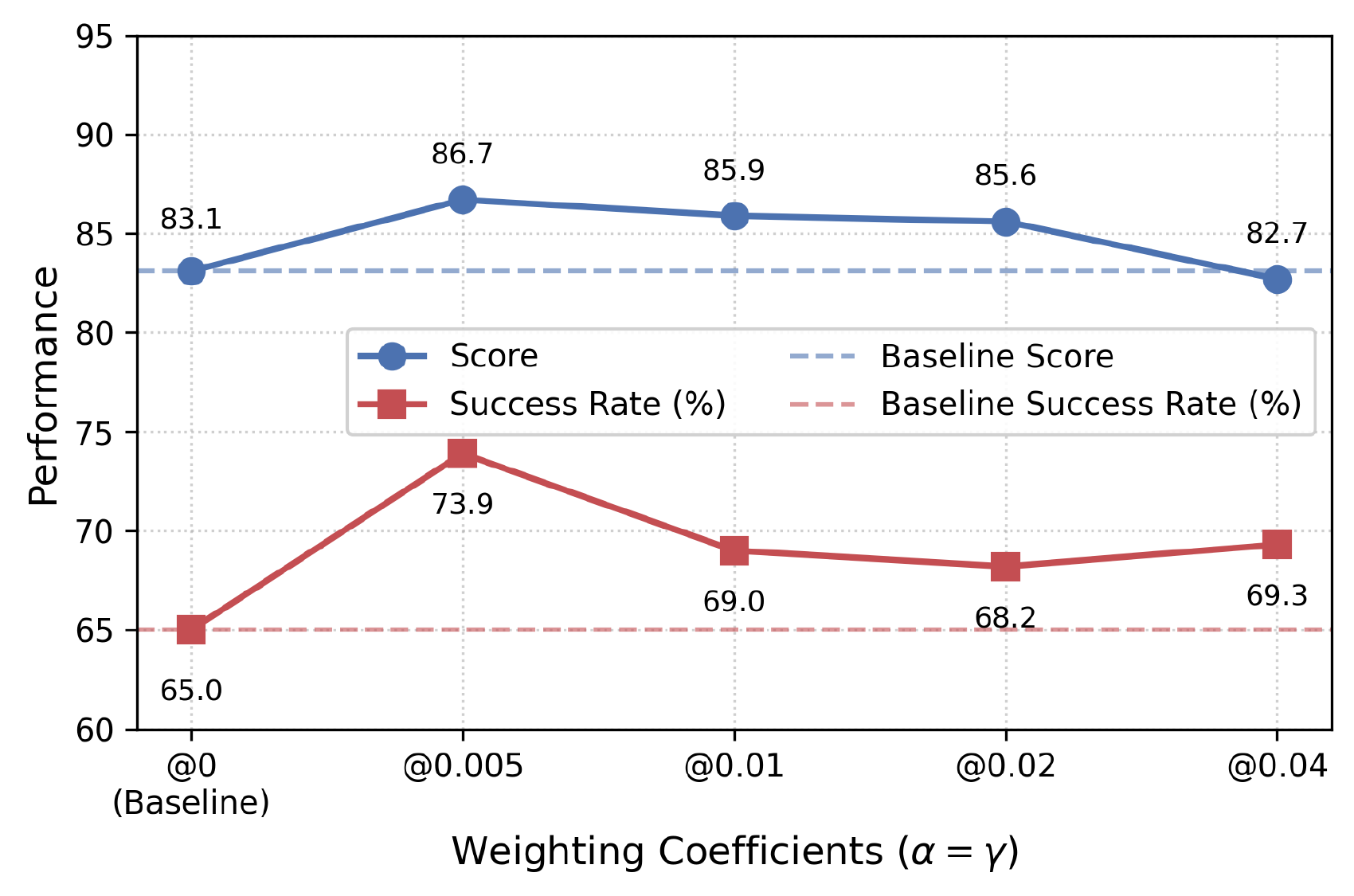}
  \caption{Sensitivity analysis of weighting coefficients $\alpha$ and $\gamma$ on WebShop, under the constraint $\alpha=\gamma$. The standard global KL penalty coefficient is fixed at $\beta=0.01$.}
  \label{fig:fig_saw}
\end{figure}

\begin{table*}[t]
\centering
\small
\begin{tabular}{lcc|cc}
\toprule
\multirow{2}{*}{Method} & \multicolumn{2}{c|}{\textbf{ALFWorld} (Outliers, \%)} & \multicolumn{2}{c}{\textbf{WebShop} (Outliers, \%)} \\ 

 & Low-Entropy & High-Entropy & Low-Entropy & High-Entropy \\ 
\midrule
\multicolumn{5}{l}{\textit{Qwen2.5-1.5B-Instruct}} \\
Base & 3.0 & 97.0 & 9.1 & 90.9 \\
GiGPO & 0.0 & 100.0 & 0.0 & 100.0 \\
STAPO & 0.0 & 100.0 & 0.0 & 100.0 \\ 
\bottomrule
\end{tabular}
\caption{Distribution of low-entropy vs. high-entropy outliers within the outlier set located by normalized entropy. The statistics are calculated based on the Qwen2.5-1.5B-Instruct model series on the test sets.}
\label{tab:outlier_composition}
\end{table*}

\begin{table*}[t]
\centering
\small 
\resizebox{\textwidth}{!}{
\begin{tabular}{lccc@{\,\,\,}c|ccc@{\,\,\,}c|c|c}
\toprule
\multirow{2}{*}{Method} & \multirow{2}{*}{Group Size} & \multicolumn{3}{c|}{\textbf{Single-Hop QA}} & \multicolumn{4}{c|}{\textbf{Multi-Hop QA}} & \multirow{2}{*}{Avg.} & \multirow{2}{*}{Time per Step} \\
& & NQ$^{\dagger}$ & TriviaQA$^{\star}$ & PopQA$^{\star}$ & HotpotQA$^{\dagger}$ & 2Wiki$^{\star}$ & MuSiQue$^{\star}$ & Bamboogle$^{\star}$ & & \\
\midrule
\multicolumn{9}{l}{\textit{Qwen2.5-7B-Instruct}} \\
\multirow{2}{*}{GiGPO} & 5 & \underline{46.4} & {64.7} & \underline{46.1} & {41.6}  & \underline{43.6} &  \textbf{18.9} & {68.9} & \underline{47.2} & \textbf{244s} \\
                       & 10 & 45.3 & \underline{65.2} & 46.0 & \underline{41.8} & 42.4 & 17.4 & \textbf{69.4} & 46.8 & 474s \\
\midrule
\textbf{STAPO} & 5 & \textbf{48.8} & \textbf{66.0} & \textbf{48.7} & \textbf{43.0}  & \textbf{45.0} &  \underline{18.1} & \textbf{69.4} & \textbf{48.4} & \underline{258s} \\

\bottomrule
\end{tabular}
}
\caption{Efficiency analysis with different group sizes on search-augmented QA tasks. Results are reported in terms of performance and training time per step. The best and second-best results are \textbf{bold} and \underline{underlined}.}
\label{tab:TE}
\end{table*}

\begin{table}[t]
\centering
\small
\begin{tabular}{lccc}
\toprule
Method & Avg.\ KL & Avg.\ Attr. & Succ.\ (\%) \\
\midrule
\multicolumn{4}{l}{\textit{Qwen2.5-1.5B-Instruct}} \\
Base & 0.40 & 0.51 & 4.1 \\
GiGPO & 0.77 & 0.60 & 86.7 \\
{STAPO} & {0.92} & {0.67} & {92.2} \\
\bottomrule
\end{tabular}
\caption{Trajectory reliance analysis on ALFWorld.}
\label{tab:appendix_causal_alf}
\end{table}

\begin{table}[t]
\centering
\small
\begin{tabular}{lccc}
\toprule
Method & Avg.\ KL & Avg.\ Attr. & Succ.\ (\%) \\
\midrule
\multicolumn{4}{l}{\textit{Qwen2.5-1.5B-Instruct}} \\
Base & 0.70 & 0.94 & 5.2 \\
GiGPO & 1.22 & 1.06 & 65.0 \\
{STAPO} & {1.61} & {1.12} & {69.0} \\
\bottomrule
\end{tabular}
\caption{Trajectory reliance analysis on WebShop.}
\label{tab:appendix_causal_ws}
\end{table}

\begin{table}[t]
\centering
\small
\begin{tabular}{lcc}
\toprule
Method & Sokoban ($6{\times}6$) & EZPoints \\
\midrule
\multicolumn{3}{l}{\textit{Qwen2.5-VL-3B-Instruct}} \\
Base & 11.7 & 3.1 \\
GRPO & 67.1 & 86.9 \\
GiGPO & 76.9 & {100.0} \\
{STAPO} & {80.5} & {100.0} \\
\bottomrule
\end{tabular}
\caption{Success rates (\%) on VLM agents.}
\label{tab:appendix_vlm}
\end{table}

\paragraph{Analysis.}
The results, illustrated in Figure~\ref{fig:fig_saw}, demonstrate that STAPO benefits significantly from the proposed trajectory-aware mechanisms within a reasonable hyperparameter range.
\begin{itemize}
    \item \textbf{Optimal Performance at Moderate Weights:} The model achieves peak performance at $\alpha=\gamma=0.005$, attaining a reward score of 86.7 and a success rate of 73.9\%. This represents a substantial improvement over the baseline, which yields a score of 83.1 and a success rate of 65.0\%, demonstrating the effectiveness of introducing trajectory-aware supervision.
    \item \textbf{Robustness and Degradation:} As the weights increase to 0.01 and 0.02, the performance decreases slightly from the peak but remains consistently superior to the baseline. However, further increasing the weights to $\alpha=\gamma=0.04$ leads to a performance drop, with the score falling to 82.7, slightly below the baseline level (83.1). This suggests that excessively high weights cause the auxiliary objectives to dominate the total loss, overshadowing the primary task reward and leading to optimization instability.
\end{itemize}
In summary, STAPO demonstrates robustness to hyperparameter variations within the range of $[0.005, 0.02]$, with $0.005$ providing the best results.

\section{Internal Composition of Outliers}
\label{sec:appendix_ICO}

While the main text demonstrates that reducing the proportion of outliers correlates with performance gains, this section investigates the internal composition of these outliers. Specifically, we analyze whether the outliers identified by our normalized entropy metric stem from high uncertainty or excessive certainty, both of which can be manifestations of trajectory neglect.

We classify outliers into two categories based on the IQR thresholds: high-entropy outliers ($H_n > Q_3 + \lambda \cdot \text{IQR}$) and low-entropy outliers ($H_n < Q_1 - \lambda \cdot \text{IQR}$). Table~\ref{tab:outlier_composition} presents the distribution of these outlier types across the Qwen2.5-1.5B-Instruct model series. We observe two key phenomena:

\paragraph{Dominance of High-Entropy Outliers.}
Across all models and datasets, the vast majority of outliers (over 90\% in the base model) are high-entropy. This indicates that trajectory neglect in long-horizon tasks primarily manifests as uncertainty or "getting lost" rather than stubborn overconfidence.

\paragraph{Elimination of Low-Entropy Outliers via RL.}
The base model does exhibit a small but noticeable fraction of low-entropy outliers (e.g., 9.1\% in WebShop), typically corresponding to repetitive loops. However, after applying RL (both GiGPO and STAPO), these low-entropy outliers are completely eliminated (0.0\%).

\section{Training Efficiency}
\label{sec:appendix_TE}
To evaluate the computational cost-effectiveness of our approach, we analyze the training efficiency on search-augmented QA tasks using Qwen2.5-7B-Instruct. As shown in Table~\ref{tab:TE}, STAPO incurs a negligible computational overhead compared to the GiGPO baseline with the same group size ($N=5$)—increasing training time per step by only $\sim$5.7\% (from 244s to 258s) primarily due to the additional forward passes required for the calculation of trajectory-aware reward and trajectory-independent penalty—yet achieves a substantial performance gain. Furthermore, STAPO demonstrates superior efficiency over brute-force scaling; simply doubling the group size for GiGPO ($N=10$) nearly doubles the training cost (474s) while paradoxically degrading performance (46.8). These results confirm that STAPO effectively trades a marginal computational cost for significant robustness and performance improvements, establishing it as a more efficient training paradigm than simple sample scaling.

\section{Trajectory Reliance Analysis}
\label{sec:appendix_causal}
On the ALFWorld and WebShop test sets, we measure (i) the average KL divergence between the policy output under the full prompt $c_t$ and under the trajectory-blind prompt $(c_t)_{\setminus\mathcal{T}}$, and (ii) the average trajectory attribution score. Higher KL indicates stronger sensitivity to trajectory context; higher attribution indicates greater use of trajectory in the model's decision. Tables~\ref{tab:appendix_causal_alf} and~\ref{tab:appendix_causal_ws} summarize the results together with success rate.

\section{Vision-Language Agent Experiments}
\label{sec:appendix_vlm}
We extend STAPO to vision-language model (VLM) agents by applying the same masking $(\cdot)_{\setminus\mathcal{T}}$ to strip the task goal and interaction history (including all previous paired visual frames and textual instructions, as well as past actions). We train {Qwen2.5-VL-3B-Instruct} with the same GiGPO baseline configuration on two visual interactive environments: \textbf{Sokoban} ($6{\times}6$), a spatial planning puzzle from image observations, and \textbf{EZPoints}, a multi-step symbolic reasoning task over visual playing cards. Table~\ref{tab:appendix_vlm} reports success rates (\%).

\clearpage
\onecolumn

\begin{figure}[t]
\centering
\resizebox{\textwidth}{!}{
\begin{tcolorbox}[colback=gray!5!white, colframe=black!75!black, 
title=Case: Neglect of Task Goal, boxrule=0.3mm, width=\textwidth, arc=3mm, auto outer arc=true]

\textbf{Task:} look at cd under the desklamp. \\

\textbf{Step 21}

\textbf{Observation:} You pick up the cd 1 from the shelf 1.

\textbf{Reasoning:} The previous step was to take the cd 1 from the shelf 1. Now, the current situation is that I have the cd 1 in my possession. I need to look at the cd 1 to understand what it is. I also need to make sure it's necessary to look at the cd 1 or if there's something else I need to do first.

\textbf{Action:} examine cd 1 \\

\textbf{Step 22}

\textbf{Observation:} There's nothing special about cd 1.

\textbf{Reasoning:} The previous action was to take cd 1 from the shelf 1. Now, there is only one cd left on the shelf 1.

\textbf{Action:} go to desk 1 \\

\textbf{Step 23}

\textbf{Observation:} You arrive at desk 1. On the desk 1, you see a cellphone 1, a creditcard 2, and a creditcard 1.

\textbf{Reasoning:} At the previous steps, I picked up and examined the CD1, and then went to the desk 1. Now, I am at step 23, at the desk 1. I see several items on the desk: a cellphone1, a creditcard2, and a creditcard1. There don't appear to be any CDs or shelves that I haven't referred to in this scenario. However, I might want to ensure I've correctly inspected the items, especially if this information is necessary for later actions.

\textbf{Action:} \underline{go to bed 1} \quad \textit{$\leftarrow$ Deviates from the goal ``desklamp''} \\

\textbf{Step 24}

\textbf{Observation:} You arrive at bed 1. On the bed 1, you see a laptop 2, a laptop 1, a pillow 1, and a teddybear 1.

\textbf{Reasoning:} The previous actions were to go to desk 1 and examine bed 1. Now, at step 24, I am at bed 1. The most recent observation is that I am at bed 1. On the bed 1, I see a laptop 2, a laptop 1, a pillow 1, and a teddybear 1.

\textbf{Action:} go to desk 2 \\

\textbf{Step 25}

\textbf{Observation:} You arrive at desk 2. On the desk 2, you see a book 1, a pen 2, a pen 1, and a pencil 2.

\textbf{Reasoning:} The previous step showed you arriving at desk 2, where you found the following objects: book 1, pen 1, pen 2, and pencil 2. Now in step 25, you should look at the cd under the desk 2. To accomplish this, you need to move the cd from your bed (desk 1) to desk 2.

\textbf{Action:} examine cd 1

\end{tcolorbox}
}
\caption{Illustration of task goal neglect in multi-turn interactions.}
\label{case:goal_neglect}
\end{figure}

\begin{figure}[t]
\centering
\resizebox{\textwidth}{!}{
\begin{tcolorbox}[colback=gray!5!white, colframe=black!75!black, 
title=Case:  Neglect of Interaction History, boxrule=0.3mm, width=\textwidth, arc=3mm, auto outer arc=true]

\textbf{Task:} look at cd under the desklamp. \\

\textbf{Step 34}

\textbf{Observation:} You arrive at shelf 1. On the shelf 1, you see a cd 2.

\textbf{Reasoning:} At step 33, you went to shelf 1, where you found a cd 2. Now, you need to look at cd 2.

\textbf{Action:} look at cd 2 \\

\textbf{Step 35}

\textbf{Observation:} Nothing happens.

\textbf{Reasoning:} The previous actions were:
1. Going to shelf 1 and finding a CD 2.
2. Looking at the CD 2.
Now, I need to decide which action to take next.

\textbf{Action:} \underline{look at cd 2} \quad \textit{$\leftarrow$ Redundant action} \\

\textbf{Step 36}

\textbf{Observation:} Nothing happens.

\textbf{Reasoning:} The previous steps showed that I have already looked at the cd 2 on the shelf 1. Now, I need to find out if there is a cd in the garbagecan, desk, shelves 2 to 5, or if I need to do something else.

\textbf{Action:} look in garbagecan 1 \\

\textbf{Step 37}

\textbf{Observation:} Nothing happens.

\textbf{Reasoning:} The user looked at the cd 2 and examine the cd 1 but there's nothing happening as no directory change or file operation instructions were provided. Also, there are some steps not taken as per the environment's logical flow. I need to plan how to reach the CD directory.

\textbf{Action:} go to desk 2 \\

\textbf{Step 38}

\textbf{Observation:} You arrive at desk 2. On the desk 2, you see a book 1, a pen 2, a pen 1, and a pencil 2.

\textbf{Reasoning:} The environment has given us two possible choices: look at cd under the desklamp. Let's check both the desk 2 and the garbagecan 1. Then, we can choose the appropriate action based on the new information.

\textbf{Action:} examine cd 1

\end{tcolorbox}
}
\caption{Illustration of interaction history neglect in multi-turn interactions.}
\label{case:history_neglect}
\end{figure}

\begin{figure}[t]
\centering
\resizebox{\textwidth}{!}{
\begin{tcolorbox}[colback=gray!5!white, colframe=black!75!black, 
title=Case: Avoiding False Positives in States with a Large Action Space, boxrule=0.3mm, width=\textwidth, arc=3mm, auto outer arc=true]

\textbf{Task:} put a cool bowl in cabinet.

\textbf{Action Type:} high-quality action (positive advantage). Identified as an {outlier} by raw Shannon entropy (IQR), but correctly as an {inlier} by normalized entropy. \\

\textbf{Step 27}

\textbf{Observation:} You arrive at cabinet 10. The cabinet 10 is open. In it, you see nothing.

\textbf{Admissible Actions}: ['close cabinet 10', 'examine cabinet 10', 'go to cabinet 1', 'go to cabinet 11', 'go to cabinet 12', 'go to cabinet 13', 'go to cabinet 2', 'go to cabinet 3', 'go to cabinet 4', 'go to cabinet 5', 'go to cabinet 6', 'go to cabinet 7', 'go to cabinet 8', 'go to cabinet 9', 'go to coffeemachine 1', 'go to countertop 1', 'go to diningtable 1', 'go to drawer 1', 'go to fridge 1', 'go to garbagecan 1', 'go to microwave 1', 'go to shelf 1', 'go to shelf 2', 'go to shelf 3', 'go to sinkbasin 1', 'go to stoveburner 1', 'go to stoveburner 2', 'go to stoveburner 3', 'go to stoveburner 4', 'go to toaster 1', 'inventory', 'look'] (\underline{Total: 32 selectable actions}).

\textbf{Reasoning:} You are at cabinet 10, the cabinet is open. There is nothing inside. The goal is to put a cool bowl in a cabinet. It seems cabinets with exposed items are more likely to contain bowls as these are often kept next to food for convenience. You should go to cabinet 1 to look for a cool bowl or any other container. Inventory of cabinets: (from observation 26) - No items found. Inventory of other locations: (additional observations) - Stoves, sink, countertops, and fridge have no cool bowls. Cabinets used in the environment typically hold food items tightly sealed. Going to other cabinets to look for bowls still seems unlikely given the cold temperature would be necessary for a bowl to be cool. The table or countertops at dining table, or any shelves or drawers within them, should also contain bowls. Placing a bowl on dining table countertops could be a viable approach too if it is a dining facility environment. However, since cabinets are often recommended for preserving shorter shelf-life items like bowls and it features free space on the countertop or shelves, reaching for a bowl is likely the best choice. Combining thoughts, going to counting table or dining table seems most likely.

\textbf{Action:} go to countertop 1

\end{tcolorbox}
}
\caption{Illustration of normalized entropy avoiding false positives in states with a large action space.}
\label{case:avoiding_false}
\end{figure}

\begin{figure}[t]
\centering
\resizebox{\textwidth}{!}{
\begin{tcolorbox}[colback=gray!5!white, colframe=black!75!black, 
title=Case: Reducing False Negatives in States with a Small Action Space, boxrule=0.3mm, width=\textwidth, arc=3mm, auto outer arc=true]

\textbf{Task:} put two spraybottle in cabinet.

\textbf{Action Type:} low-quality action (negative advantage). Identified as an {inlier} by raw Shannon entropy (IQR), but correctly as an {outlier} by normalized entropy. \\

\textbf{Step 43}

\textbf{Observation:} The cabinet 2 is open. In it, you see nothing.

\textbf{Admissible Actions}: ['close cabinet 2', 'examine cabinet 2', 'go to bathtubbasin 1', 'go to cabinet 1', 'go to cabinet 3', 'go to cabinet 4', 'go to countertop 1', 'go to garbagecan 1', 'go to handtowelholder 1', 'go to sinkbasin 1', 'go to sinkbasin 2', 'go to toilet 1', 'go to toiletpaperhanger 1', 'go to towelholder 1', 'inventory', 'look'] (\underline{Total: 16 selectable actions}).

\textbf{Reasoning:} In the current situation, we have taken 42 steps, but we have not seen the two spraybottles placed anywhere. We have checked cabinet 2, but it is empty. To place the spraybottles, we need to clear the space inside the cabinet first, and then find the spraybottles.

\textbf{Action:} close cabinet 2

\end{tcolorbox}
}
\caption{Illustration of normalized entropy reducing false negatives in states with a small action space.}
\label{case:reducing_false}
\end{figure}

\begin{figure}[t]
\centering
\resizebox{\textwidth}{!}{
\begin{tcolorbox}[colback=gray!5!white, colframe=black!75!black, 
title=Complete Prompt Template for ALFWorld, boxrule=0.3mm, width=\textwidth, arc=3mm, auto outer arc=true]
You are an expert agent operating in the ALFRED embodied Environment. Your task is to: \{task\_description\}. Prior to this step, you have already taken \{step\_count\} step(s). Below are the most recent \{history\_length\} observations and the corresponding actions you took: \{action\_history\}. You are now at step \{current\_step\} and your current observation is: \{current\_observation\}. Your admissible actions of the current situation are: [\{admissible\_actions\}]. \\

Now it's your turn to take an action. You should first reason step-by-step about the current situation. This reasoning process MUST be enclosed within <think> </think> tags. Once you've finished your reasoning, you should choose an admissible action for current step and present it within <action> </action> tags.
\end{tcolorbox}
}
\caption{The complete prompt template of ALFWorld agents.}
\label{prompt:alfworld_complete}
\end{figure}

\begin{figure}[t]
\centering
\resizebox{\textwidth}{!}{
\begin{tcolorbox}[colback=gray!5!white, colframe=black!75!black, 
title=Trajectory-Blind Prompt Template for ALFWorld, boxrule=0.3mm, width=\textwidth, arc=3mm, auto outer arc=true]
You are an expert agent operating in the ALFRED Embodied Environment. You are now at step \{current\_step\} and your current observation is: \{current\_observation\}. Your admissible actions of the current situation are: [\{admissible\_actions\}]. \\

Now it's your turn to take an action. You should first reason step-by-step about the current situation. This reasoning process MUST be enclosed within <think> </think> tags. Once you've finished your reasoning, you should choose an admissible action for current step and present it within <action> </action> tags.
\end{tcolorbox}
}
\caption{The trajectory-blind prompt template of ALFWorld agents.}
\label{prompt:alfworld_blind}
\end{figure}

\begin{figure}[t]
\centering
\resizebox{\textwidth}{!}{
\begin{tcolorbox}[colback=gray!5!white, colframe=black!75!black, 
title=Complete Prompt Template for WebShop, boxrule=0.3mm, width=\textwidth, arc=3mm, auto outer arc=true]
You are an expert autonomous agent operating in the WebShop e-commerce environment. Your task is to: \{task\_description\}. Prior to this step, you have already taken \{step\_count\} step(s). Below are the most recent \{history\_length\} observations and the corresponding actions you took: \{action\_history\}. You are now at step \{current\_step\} and your current observation is: \{current\_observation\}. Your admissible actions of the current situation are: [\{available\_actions\}]. \\

Now it's your turn to take one action for the current step. You should first reason step-by-step about the current situation, then think carefully which admissible action best advances the shopping goal. This reasoning process MUST be enclosed within <think> </think> tags. Once you've finished your reasoning, you should choose an admissible action for current step and present it within <action> </action> tags.
\end{tcolorbox}
}
\caption{The complete prompt template of WebShop agents.}
\label{prompt:webshop_complete}
\end{figure}

\begin{figure}[t]
\centering
\resizebox{\textwidth}{!}{
\begin{tcolorbox}[colback=gray!5!white, colframe=black!75!black, 
title=Trajectory-Blind Prompt Template for WebShop, boxrule=0.3mm, width=\textwidth, arc=3mm, auto outer arc=true]
You are an expert autonomous agent operating in the WebShop e-commerce environment. You are now at step \{current\_step\} and your current observation is \{current\_observation\}. Your admissible actions of the current situation are: [\{available\_actions\}]. \\

Now it's your turn to take one action for the current step. You should first reason step-by-step about the current situation, then think carefully which admissible action best advances the shopping goal. This reasoning process MUST be enclosed within <think> </think> tags. Once you've finished your reasoning, you should choose an admissible action for current step and present it within <action> </action> tags.
\end{tcolorbox}
}
\caption{The trajectory-blind prompt template of WebShop agents.}
\label{prompt:webshop_blind}
\end{figure}

\begin{figure}[t]
\centering
\resizebox{\textwidth}{!}{
\begin{tcolorbox}[colback=gray!5!white, colframe=black!75!black, 
title=Complete Prompt Template for Search, boxrule=0.3mm, width=\textwidth, arc=3mm, auto outer arc=true]
You are an expert agent tasked with answering the given question step-by-step. Your question: \{task\_description\}. Prior to this step, you have already taken \{step\_count\} step(s). Below is the interaction history where <search> </search> wrapped your past search queries and <information> </information> wrapped the corresponding search results returned by the external search engine. History: \{memory\_context\}. \\

Now it's your turn to respond for the current step. You should first conduct reasoning process. This process MUST be enclosed within <think> </think> tags. After completing your reasoning, choose only one of the following actions (do not perform both):

(1) If you find you lack some knowledge, you can call a search engine to get more external information using format: <search> your query </search>.

(2) If you have enough knowledge to answer the question confidently, provide your final answer within <answer> </answer> tags, without detailed illustrations. For example, <answer>Beijing</answer>.
\end{tcolorbox}
}
\caption{The complete prompt template of Search agents.}
\label{prompt:search_complete}
\end{figure}

\begin{figure}[t]
\centering
\resizebox{\textwidth}{!}{
\begin{tcolorbox}[colback=gray!5!white, colframe=black!75!black, 
title=Trajectory-Blind Prompt Template for Search, boxrule=0.3mm, width=\textwidth, arc=3mm, auto outer arc=true]
You are an expert agent tasked with answering the given question step-by-step. \\

Now it's your turn to respond for the current step. You should first conduct reasoning process. This process MUST be enclosed within <think> </think> tags. After completing your reasoning, choose only one of the following actions (do not perform both):

(1) If you find you lack some knowledge, you can call a search engine to get more external information using format: <search> your query </search>.

(2) If you have enough knowledge to answer the question confidently, provide your final answer within <answer> </answer> tags, without detailed illustrations. For example, <answer>Beijing</answer>.
\end{tcolorbox}
}
\caption{The trajectory-blind prompt template of Search agents.}
\label{prompt:search_blind}
\end{figure}

\end{document}